\documentclass[review]{elsarticle}

\usepackage{lineno,hyperref}
\modulolinenumbers[5]

\usepackage{booktabs}       
\usepackage{amsfonts}       
\usepackage{nicefrac}       
\usepackage{microtype}      

\usepackage{url}
\usepackage{amsmath,amssymb} 
\usepackage{color}
\usepackage{graphicx}
\usepackage{algorithm}
\usepackage{algorithmicx}
\usepackage{algpseudocode}
\usepackage{algpascal}

\usepackage{mathrsfs}
\usepackage{soul}
\usepackage{multirow}
\usepackage{comment}
\usepackage{xfrac}
\usepackage{rotating}
\usepackage{hhline}
\usepackage[table]{xcolor}
\usepackage{mathtools}
\usepackage{wrapfig}

\usepackage{subfig}
\usepackage{pgfplots}
\usepackage{pgfplotstable}
\usepackage{filecontents,pgfplots}
\usepackage{tikz}
\usetikzlibrary{arrows,calc,shapes,snakes,positioning,matrix,arrows,decorations.pathmorphing,decorations.text}
\newcommand{\AL}{\boldsymbol{\alpha}}
\usepackage{textcomp}
\usepackage{sci}

\newcommand{\rev}[1]{{#1}}

\journal{MEDIA}

\begin{document}

\begin{frontmatter}

\title{Longitudinal Self-Supervised Learning}

\author[stanford_psy]{Qingyu Zhao\corref{contrib}}
\author[stanford_ee]{Zixuan Liu\corref{contrib}}
\author[stanford_psy,stanford_cs]{Ehsan Adeli}
\author[stanford_psy,sri]{Kilian M. Pohl\corref{mycorrespondingauthor}}

\cortext[contrib]{Authors contributed equally}
\cortext[mycorrespondingauthor]{Corresponding author}\ead{kilian.pohl@stanford.edu}

\address[stanford_psy]{Department of Psychiatry \& Behavioral Sciences, Stanford University, Stanford, CA 94305}
\address[stanford_ee]{Department of Electrical Engineering, Stanford University, Stanford, CA 94305}
\address[stanford_cs]{Department of Computer Science, Stanford University, Stanford, CA 94305}
\address[sri]{Center for Biomedical Sciences, SRI International, Menlo Park, CA 94205}

\begin{abstract}
  
   Machine learning analysis of longitudinal neuroimaging data is typically based on supervised learning, which requires large number of ground-truth labels to be informative. 
   As ground-truth labels are often missing or expensive to obtain in neuroscience,  we avoid them in our analysis by combing factor disentanglement with self-supervised learning to identify changes and consistencies across the multiple MRIs acquired of each individual over time. Specifically, we propose a new definition of disentanglement by formulating a multivariate mapping between factors  (e.g., \textit{brain age}) associated with an MRI and a latent image representation. Then, factors that evolve across acquisitions of longitudinal sequences are disentangled from that mapping by self-supervised learning in such a way that changes in a single factor induce change along one direction in the representation space. We implement this model, named Longitudinal Self-Supervised Learning (LSSL), via a standard autoencoding structure with a cosine loss to disentangle brain age from the image representation. We apply LSSL to two longitudinal neuroimaging studies to highlight its strength in extracting the brain-age information from MRI and revealing informative characteristics associated with neurodegenerative and neuropsychological disorders. Moreover, the representations learned by LSSL facilitate supervised classification by recording faster convergence and higher (or similar) prediction accuracy compared to several other representation learning techniques.

\end{abstract}

\begin{keyword}
Longitudinal Neuroimaging \sep Self-supervised Learning \sep Factor Disentanglement \sep Brain Age
\end{keyword}

\end{frontmatter}


\section{Introduction} 

A longitudinal study repeatedly measures the same variable to track a specific group of individuals over a period of time \citep{Caruana2015}. For example, longitudinal neuroimaging studies are often used to evaluate the impact of age on the brain \citep{Dennis2008}, the relationship between risk factors and development of disease \citep{adni2008}, and the outcomes of treatments over time \citep{Liu2010}. A critical component of longitudinal studies is to apply data analysis approaches properly modelling the complex correlations underlying the repeated measures, which are often characterized by a mixture of inter-subject variance and intra-subject dependency. Popular analysis approaches are mixed-effect models \citep{bernal2013statistical} and analysis of variance (ANOVA) \citep{anova2016}, which can inspect the influence of key factors (e.g., age or disease) on individual imaging measurements (e.g., cortical thickness of regions of interest) extracted from the longitudinal Magnetic Resonance images (MRIs). However, this type of univariate analysis ignores the high-dimensional relations across multiple brain regions \citep{Habeck2010cell}. With recent advances in deep learning, this limitation can be largely resolved by data-driven supervised learning, i.e.,  by training models to predict the value of a set of factors (e.g., age or diagnosis group) for each subject directly from their raw images \citep{aghili2018predictive}.

One limitation of supervised learning is that the training requires large amount of data with accurate labels, which is infeasible for some neuroimaging applications. For example, prior studies have trained models to predict subjects' age from their structural MRIs to understand the effect of aging on brain morphometry \citep{zhao2019variational,Smith2020eLife}. However, such age-based supervision can be sub-optimal as the chronological age does not always reflect the true condition of aging in the brain \citep{Steffener2016}. To resolve this issue, we propose here a novel learning scheme that replaces the direct supervision with \textit{self-supervision} in the context of repeated measures. In general, self-supervised learning \citep{Kolesnikov2019} aims to automatically generate \textit{supervisory signal} by exploring similarity and dissimilarity relations across samples through the learning process of their representations. This concept aligns with the longitudinal design, in which each subject serves as their own `comparison' with respect to the change over time. Therefore, the repeated measures could fully leverage the structured inter-relation across time points to learn time-dependent representations of MRIs.


Another concept in representation learning that can be related to the longitudinal design is \textit{factor disentanglement} \citep{Tschannen2018}. The key intuition behind disentanglement is that real-world data have low-dimensional representations that are controlled by distinct and interpretable factors \citep{higgins2018definition}. Changing a single factor should thus lead to a change in a single dimension in the representation space. According to a recent publication \citep{locatello2018challenging}, factor disentanglement cannot be stringently formulated for cross-sectional data without explicitly supervision. 
However, in a longitudinal study, where each individual is repeatedly examined over time, all time-independent factors are fixed so that the impact of time-related factors, such as age, can be effectively revealed on the observations. In other words, the study design of repeated measures fits perfectly to `supervising' the disentanglement of the evolving factor across the measures.

Inspired by these affinities, we propose a representation learning strategy, named Longitudinal Self-Supervised Learning (LSSL), to investigate the impact of aging and neuropsychological disorders on the brain within the context of longitudinal MRIs. Thanks to the repeated measure design in longitudinal studies, we separately define the space of factors and the representation space of MRIs and formulate disentanglement in the context of multivariate mapping between the two spaces. The factor of brain age is then disentangled by self-supervision on the temporal order between the multiple MRIs for each subject. The training uses this ordinal information to encourage the change of subject-specific representations to follow a common developmental direction. This objective is condensed to optimizing a combination of image reconstruction loss of a standard autoencoder \citep{Chen2017ae} and a simple cosine loss in the representation space. 

We test our model on \rev{a synthetic dataset and two longitudinal MR data sets: one investigates the impact of Alzheimer's Disease (AD) on the brain and the other one of alcohol dependence}. While the training does not rely on any ground-truth age nor diagnostic labels, LSSL successfully disentangles a factor in the representation space linked to brain age, which is superior than the chronological age in characterizing the health condition of the brain. In addition, LSSL successfully reveals accelerated aging effects of AD and alcohol dependence compared to the control cohort. When performing a downstream task of predicting diagnosis labels of subjects, the representations and pre-trained encoder learned by our model result in faster convergence and more (or equivalently) accurate classification accuracy compared to several commonly used state-of-the-art unsupervised or self-supervised representation learning strategies.

\section{Related Work}

\textbf{Longitudinal Neuroimaging Studies.} Deep learning models applied to longitudinal neuroimaging studies are largely based on supervised learning, i.e., by training classification models based on image time series using regular recurrent networks \citep{lipton2015learning,santeramo2018longitudinal,cui2019rnn,ghazi2019training}. Based on these models, longitudinal pooling is proposed to augment the learned representation of each time point with information gathered from images of other time points in the series \citep{ouyang2020recurrent}. Other methods for explicitly exploiting dependencies within the intra-subject series are generally based on parameterizing the trajectories of representations in the latent space, such as using Mixed Effect Models \citep{Xiong2019,Louis2019}. 

\textbf{Brain-age Analysis.} The biological `brain age' is a marker quantifying the health condition of the brain and correlates with mental and physical fitness \citep{Steffener2016}. To predict the brain age from structural MRIs, recent publications \citep{cole2016neuroimage,Smith2020eLife,zhao2019variational,Elliott2019,Kaufmann2019} proposed to train supervised learning models on a cohort of healthy subjects with the prediction target being their chronological age. Such models can be built on imaging measurements \citep{Smith2020eLife} or raw 3D images \citep{zhao2019variational} and in both cross-sectional \citep{Smith2020eLife} and longitudinal settings \citep{Elliott2019}. After training, the model can be applied to both healthy and diseased subjects, and the difference between the predicted brain age and chronological age can be regarded as a phenotype related to the impact of the brain disorder on normal aging \citep{Kaufmann2019}. Despite the meaningful findings revealed in these studies, one common pitfall is the underlying assumption that the chronological age exactly corresponds to the brain age, which is challenged by several recent studies that show brain aging is highly heterogeneous even within the healthy populations \citep{Franke2019} .

\textbf{Self-Supervised Learning.} To tackle the problem of missing or expensive-to-obtain ground-truth labels required by supervised learning, self-supervised methods learn representations by training models in tasks that do not require explicit annotations or labels \citep{jing2019selfsupervised}, such as image inpainting \citep{zhang2016colorful}, colorization \citep{larsson2017colorization}, super-resolution \citep{dong2014learning,johnson2016perceptual}, and predicting spatial relationships between two image patches from the same image \citep{noroozi2016unsupervised,sabokrou2019self}. Beyond these within-sample learning schemes, contrastive learning is a self-supervised approach that explicitly models the relationships across samples, such as by distinguishing between similar and dissimilar images \citep{simclr}, modeling temporal dependency across time \citep{oord2018representation}, and estimating depth from stereo images \citep{pillai2019superdepth}. Once trained, the resulting representations can be embedded into supervised learning tasks, such as multi-task and cross-domain feature learning \citep{doersch2017multi,ren2018cross}, which result in a more efficient training with respect to labelled data and computational resources than training the supervised models from scratch \citep{simclr,moco,Kolesnikov2019}.


\textbf{Factor Disentanglement.} While there is no consensus on the mathematical definition of \textit{disentanglement}, conceptually, a representation is considered disentangled if changes along one dimension of the representation are explained by a specific factor of variation (e.g., age) while being relatively invariant to other factors (e.g. gender, ethnicity) \citep{higgins2018definition}. Most existing works formulate this notion from a statistical perspective by pursuing statistical independence among random variables in the latent space (factorizable latent representations \citep{Tschannen2018,higgins2018definition}). Therefore, state-of-the-art approaches for unsupervised disentanglement learning are based on a Variational Autoencoder (VAEs) structure, which aims to learn a factorizable posterior from the marginal distribution of the observed data \citep{Higgins2017betaVAELB,chen2018isolating,kim2018disentangling,zhao2017infovae}. Despite promising results, the study by \cite{locatello2018challenging} challenges the theoretical validity of this idea. They point out that given any marginal distribution of the observed data, there exists an infinite number of generative processes from either disentangled or fully entangled latent representations. Therefore, a true factor disentanglement \rev{requires supervision, for which we propose to perform self-supervision on repeated measures}.

\section{Longitudinal Self-Supervised Learning}
\begin{figure}[!t]
    \centering
    \includegraphics[width=\linewidth]{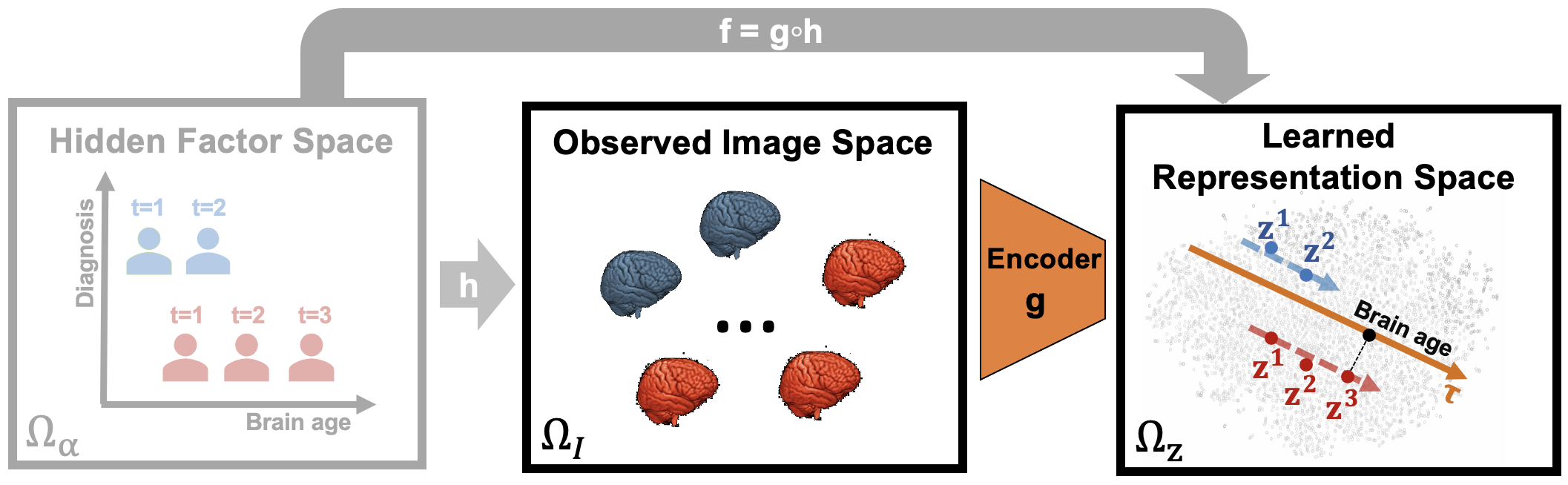} 
    \caption{\rev{Longitudinal Self-Supervised Learning (LSSL) aims to learn representations from observed images, which are assumed to be generated from a set of hidden factors}. In this example, the variation of the repeated measures of two subjects (blue and red, $t$ encodes the order of visits) is assumed to relate to an increase in brain age. LSSL then disentangles a 1D direction $\boldsymbol{\tau}$ linked to brain age from the representation space such that the developmental trajectories of subject-specific representations $\boldsymbol{z}^t$ are colinear with $\boldsymbol{\tau}$. }
    \label{fig:lssl}
\end{figure}

We now first provide a new perspective of factor disentanglement \rev{by defining a multivariate mapping from a hidden factor space underlying the observed images to a representation space learned from those images (see Fig. \ref{fig:lssl}). This setup motivates us to define a novel self-supervised objective function that does not depend on the statistical property of the disentangled factor.} We then employ this formulation in the context of the repeated measures design to disentangle brain age from longitudinal MRI data.

\subsection{Disentangled Mapping from Factor to Representation Space.}
Current unsupervised learning frameworks often define the concept of \textit{disentanglement} as the observed data having factorizable distributions in the representation space, such that each representation variable corresponds to a real-world factor that is statistically independent from other variables \citep{Higgins2017betaVAELB}. This learning objective is impractical because factors in real world datasets are not necessarily statistically independent. For example, brain morphology is influenced by both gender and brain size, two highly correlated factors. Thus, deriving a 2D representation of the underlying distribution could yield two statistically independent directions (such as with Principal or Independent Component Analysis), but the meaning of each direction with respect to real-world factors would be unclear. To resolve this issue, we step away from the statistical formulation and approach the disentanglement problem from a pure algebraic perspective. Specifically, we propose to explicitly separate the concepts of the \textit{space of factors} from the representation space. We do so by assuming images are generated by factors, and hence can be reduced to a low-dimensional representation (see Fig. \ref{fig:lssl}). We then are interested in the deterministic multivariate mapping function between the two spaces as a means of disentanglement.

To mathematically formulate such mappings, we denote $\Omega_\alpha=\mathbb{R}^M$ as the \textit{factor space}, and $\Omega_I=\mathbb{R}^P$ the \textit{image space}. We assume each image $\boldsymbol{I} \in \Omega_I$ can be fully determined based on $M$ factors $\AL:=[\alpha_1,...,\alpha_M] \in \Omega_\alpha$ through a differentiable generative process $\boldsymbol{I} = \h(\AL)$.  Further, we aim to learn a differentiable encoder $\g(\cdot)$  that reduces the image $\boldsymbol{I}$ to a $K$ dimensional representation $\boldsymbol{z}:=[z_1,...,z_K]$ in the \textit{representation space} $\Omega_z$, i.e., $\boldsymbol{z}=\g(\boldsymbol{I}) \in \Omega_z$ and $z_i=g_i(\boldsymbol{I})$. We then define $\f:\Omega_\alpha \rightarrow \Omega_z$ as the composite multivariate-to-multivariate mapping $\f:=\g \circ \h$, where $\f=[f_1,...,f_K]$ and each $f_i$ is a differentiable multivariate-to-univariate mapping with $z_i := f_i(\AL)$ for $i\in [1,K]$. 

Without loss of generality, we assume that $z_1$ is linked to the first factor $\alpha_1$. Let $\hat{\AL}:=[\alpha_2,...,\alpha_M]$, we then consider the factor $\alpha_1$ as disentangled within the representation $\boldsymbol{z}$ if $\f$ can be factorized such that 
\begin{align}
    z_1 &= f_1(\alpha_1) \mbox{ with } f_1:\mathbb{R} \rightarrow \mathbb{R} \mbox{ being strictly increasing, and } \nonumber \\
    z_i &= f_i(\hat{\AL})\mbox{ for $i>1$}.
    \label{eq:factorization}
\end{align}
    
In other words, disentanglement is achieved when 1) $z_1$ is solely dependent on $\alpha_1$ (the monotonicity of $f_1$ ensures the mapping to be bijective, i.e., without loss of information, and preserve ordinal information of factor $z_1$); and 2) the remaining representation in $\boldsymbol{z}$ is solely dependent on factors other than $\alpha_1$.

\subsection{Self-Supervised Disentanglement.} 
In many applications, the only available data are a set of images $\boldsymbol{I}$. The underlying generative process, including the mapping function $\h$, dimension of factor space $M$, and values of $\boldsymbol{\alpha}$ are hidden from observation, so training disentangled representation with respect to a specific factor can become extremely challenging. However, in situations where each training sample has multiple images measured with respect to different values of a specific factor, we can leverage self-supervision to achieve disentanglement. To show this, the factorization of $\f$ in Eq. \eqref{eq:factorization} can be transformed to the following conditions:
\begin{equation}
    \frac{\partial f_1}{\partial \alpha_1}>0\mbox{, and } \frac{\partial f_i}{\partial \alpha_1}=0 \mbox{, } \frac{\partial f_1}{\partial \alpha_i}=0 \mbox{ for $i>1$.}
    \label{eq:factorization_derivative}
\end{equation}

Based on the chain rule, the partial derivative of $f_i$ with respect to $\alpha_j$ can be further transformed to a directional derivative with respect to the vector $\boldsymbol{u}_j$:
\begin{align}
    \centering
    \boldsymbol{u}_j &= [\frac{\partial h_1}{\partial \alpha_j},...,\frac{\partial h_P}{\partial \alpha_j}],\\
    \frac{\partial f_i}{\partial \alpha_j} &= \sum_{p=1}^P \frac{\partial f_i}{\partial h_p}\frac{\partial h_p}{\partial \alpha_j}=\nabla_{\boldsymbol{u}_j} g_i(\boldsymbol{I}).
    \label{eq:directional_derivative}
\end{align}

Eq. \eqref{eq:directional_derivative} translates the problem setup on $\f$ and $\boldsymbol{\alpha}$ to a setup with respect to $\g$ and $\boldsymbol{I}$. Specifically, let $\boldsymbol{u}_j$ be the change in the image space after perturbing the value of $\alpha_j$ by $\epsilon$ and  $\boldsymbol{I}':= \boldsymbol{I}+\boldsymbol{u}_j$, then the corresponding change in $z_i$ can be defined as  
\begin{equation}
\frac{\partial f_i}{\partial \alpha_j} \approx \frac{g_i(\boldsymbol{I}')-g_i(\boldsymbol{I})}{\epsilon}.
\end{equation}
As such, the disentanglement defined by Eq. \eqref{eq:factorization_derivative} is achieved if and only if

\begin{enumerate}
    \item[] \textit{Condition 1: Upon perturbation of $\alpha_1$, $\g(\boldsymbol{I}')-\g(\boldsymbol{I})$ is co-linear with $[1,0,...,0]$;}
    \item[] \textit{Condition 2: Upon perturbation of $\alpha_i$ for $i>1$, $g_1(\boldsymbol{I}')-g_1(\boldsymbol{I})=0$.}
\end{enumerate}

\rev{In other words, to disentangle $\alpha_1$ from the representation space, one should optimize for an encoder $\g$ such that the two conditions apply to the representations of all pairs of images.}
This objective is self-supervised as we only need to pair the images but not to provide the true values of the factor.

\subsection{Studying Brain Aging via Longitudinal MRIs}
As mentioned, brain age characterizes the apparent health condition of the brain but not necessarily equals chronological age. For example, a patient with neurodegenerative disease can have a higher brain age than a healthy subject albeit they have the same chronological age \citep{Elliott2019,Kaufmann2019}. This section describes how to leverage the prior self-supervised learning model to disentangle brain age from longitudinal structural MRIs. 

To do so, we assume that 
brain age is the dominant factor that changes the brain morphology of an individual across the longitudinal scans while the other genotypic and demographic factors (such as gender and ethnicity) are static over time. Based on this assumption, \textit{Condition 2} is omitted from the following analysis as one can not `perturb' those static factors to examine their influence on the image representations in a longitudinal design. As such, we now present the Longitudinal Self-Supervised Learning (LSSL) algorithm \rev{that performs disentanglement guided by \it{Condition 1}.} 


Let $\mathcal{I}$ be the collection of all MR images and $\mathcal{S}$ be the set of subject-specific image pairs; i.e., $\mathcal{S}$ contains all $\left<\boldsymbol{I}^t, \boldsymbol{I}^s\right>$ such that $\boldsymbol{I}^t$ and $\boldsymbol{I}^s$ are from the same subject with $\boldsymbol{I}^t$ scanned before $\boldsymbol{I}^s$. \rev{To apply  \textit{Condition 1} to disentangle brain age from an image pair, we first assume that the small increase of brain age between the two time points corresponds to the perturbation of $\alpha_1$. Next, we relax the colinearity constraint of \textit{Condition 1} that the disentangled direction has to align with the first natural coordinate axe in the representation space (i.e., [1,0,...,0]).} Instead, we parameterize the direction linked to brain age as a free-form 1D unit vector $\boldsymbol{\tau}\in \Omega_z$ that can be jointly learned during training (Fig. \ref{fig:lssl}). This strategy is motivated by the findings of \cite{rolinek2018variational}, which suggest that the encoder network by itself does not have the capacity to model arbitrary rotations of the representation space. 

\rev{In doing so, the brain age associated with the two images, $\psi^t$ and $\psi^s$, can be defined as the projections of the corresponding image representations to $\boldsymbol{\tau}$}, e.g., $\psi^t=\g(\boldsymbol{I}^t)^\top\boldsymbol{\tau}={\boldsymbol{z}^t}^\top\boldsymbol{\tau}$ (Fig. \ref{fig:lssl}). Then, ensuring \textit{Condition 1} while preserving $\psi^s > \psi^t$ is equivalent to enforcing $\cos\left(\g(\boldsymbol{I}^s)-\g(\boldsymbol{I}^t),\boldsymbol{\tau}\right)=1$, i.e., a zero-angle between $\boldsymbol{\tau}$ and the direction of progression in the representation space. In other words, while the location of $\boldsymbol{z}^t$ can be arbitrary in the representation space, the change of image representation between time points $\boldsymbol{z}^s-\boldsymbol{z}^t$ is only allowed in the $\boldsymbol{\tau}$ direction. 

To learn an encoder $\g$ that satisfies the cosine constraint, we train a standard autoencoder that models $\g$ as a neural network with parameters $\theta$ and simultaneously determines the parameters $\phi$ of a decoder network $\d$ to reconstruct the input image from the encoded representation. In doing so, the learned representations encode all morphological information of the brain beyond the single factor of interest. To impose the cosine constraint in the autoencoder, we propose to add a cosine loss for each image pair to the standard Mean-Squared Error loss MSE($\cdot$,$\cdot$) of the autoencoder \rev{as a soft constraint}, i.e, 
\begin{equation}
    \min_{\theta,\phi,\boldsymbol{\tau}}  \sum_{\boldsymbol{I} \in \mathcal{I}} \mbox{MSE}\left(\boldsymbol{I},\d\left( \g \left(\boldsymbol{I};\theta \right);\phi\right)\right) 
     - \lambda \sum_{\left<\boldsymbol{I}^t,\boldsymbol{I}^s\right> \in \mathcal{S}} \cos\left(\g(\boldsymbol{I}^s;\theta)-\g(\boldsymbol{I}^t;\theta),\boldsymbol{\tau}\right),
    \label{eq:objective_function}
\end{equation}
with $\lambda$ being the parameter weighting the two terms. As a result, the objective function encourages the encoder to learn the low-dimensional representation of images while encouraging the development of brain representation of all subjects to be colinear with $\boldsymbol{\tau}$.

\section{Experiments}
\subsection{Experimental Setup}
\textbf{Datasets.}
\rev{We evaluated the proposed LSSL on a synthetic and two longitudinal neuroimaging datasets. The synthetic dataset contained 512 subjects. Each subject consisted of an image pair $\boldsymbol{I}^1$ and $\boldsymbol{I}^2$, whose difference was regarded as the developmental effect over time. Either image in the pair contained four Gaussian patterns (Fig. \ref{fig:synthetic_results}(a)). The magnitude of the two diagonal Gaussians were randomly sampled from a uniform distribution $\mathcal{U}(1,6)$ and was kept the same for the pair. The magnitude of the two off-diagonal Gaussians simulated the `brain age', which was sampled from $\psi \sim \mathcal{U}(1,4)$ in the first image and set to $\psi+\Delta \psi$ for the second image with $\Delta \psi \sim \mathcal{U}(0.1,2)$ being the age-related increase in magnitude between the two images (Fig. \ref{fig:synthetic_results}(a)). Gaussian noise of SNR=8 was added to each image. Training on the 512 image pairs, the goal of the synthetic experiment was to show that LSSL can disentangle a direction $\boldsymbol{\tau}$ in the latent space encoding the off-diagonal developmental pattern and that the estimated brain age $\psi$ (projections along $\boldsymbol{\tau}$) accurately correlates with the ground-truth $\psi$.

Next, we evaluated LSSL on the $2,641$} structural MRIs of 811 subjects from the Alzheimer’s Disease Neuroimaging Initiative (ADNI1\footnote{Data publically available at \url{http://adni.loni.usc.edu/}}). The dataset consisted of 229 normal control subjects (age: 76 $\pm$ 5.0 years), 397 subjects diagnosed with Mild Cognitive Impairment (74.9 $\pm$ 7.4 years), and 185 subjects with Alzheimer’s Disease (75.3 $\pm$ 7.6 years). \rev{The longitudinal MRI of each subject was composed of up to 8 scans (acquired within a 4 year study period) that we were able to successfully preprocess}. In line with our prior studies \citep{adeli2020deep,zhaoadeli2020cf-net}, the preprocessing consisted of denoising, bias field correction, skull striping, affine registration to a template, re-scaling to a 64 $\times$ 64 $\times$ 64 volume, and transforming image intensities within the brainmask to z-scores. \rev{We constructed 3,141 image pairs based on the criteria that each pair belonged to the same subject and had at least one year interval in scan time. }

Another neurological disorder known to accelerate brain aging is alcohol dependence, which  can cause gradual deterioration in the gray and white matter tissue \citep{Pfefferbaum14,Zahr2017}. Therefore, the second \rev{MRI} dataset was comprised of \rev{1,499} T1-weighted MRIs of 274 Normal Controls (age: 47.3 $\pm$ 17.6) and 329 patients diagnosed with alcohol dependence (age: 49.3 $\pm$ 10.5) according to the DSM-IV criteria \citep{DSM-IV} (referred to as the alcohol data set). \rev{74 participants of the alcoholic group were also human immunodeficiency virus (HIV) positive. Each subject had up to 13 longitudinal scans.} \rev{1,071 image pairs were constructed from this dataset based on the above one-year-interval criterion.} The study was approved by the institutional review boards of Stanford University School of Medicine and SRI International. All MRIs were pre-processed using the prior pipeline. 

\begin{figure}[!t]
    \centering
    \includegraphics[width=\linewidth]{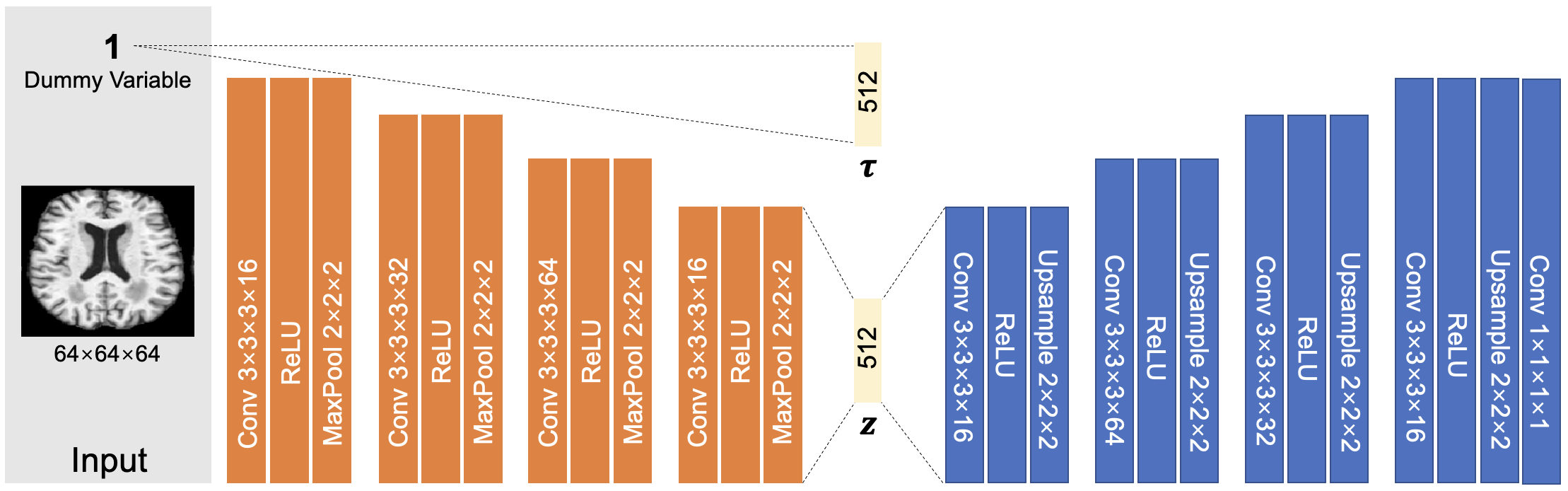} 
    \caption{\rev{Network structure of LSSL. Orange blocks correspond to the encoder and blue blocks to the  decoder networks.} }
    \label{fig:network}
\end{figure}

\textbf{Implementation of LSSL.}
\rev{As LSSL is a formulation on latent representations (which is not specific to any design of the decoder $\textbf{d}$ and encoder $\textbf{g}$), we constructed rather simplistic and standard autoencoders so that findings revealed herein are likely to generalize to more advanced autoencoder structures. For the synthetic dataset, we designed the encoder of LSSL as 3 stacks of $3\times 3$ convolution with feature dimension $\{2,4,8\}$, tanh activation, and max-pooling layers, which resulted in a 16 dimensional representation.} The decoder employed a reverse structure of the encoder. \rev{The unit vector $\boldsymbol{\tau}$ was embedded in the network as the output of a fully connected layer applied to a dummy input scalar (Fig. \ref{fig:network}), which made $\boldsymbol{\tau}$ a global variable and independent of the specific input image.} \rev{For the MRI datasets, the encoder was 4 stacks of } $3\times 3 \times 3$ convolution with dimension $\{16, 32, 64, 16\}$, ReLU activation, and max-pooling layers (Fig. \ref{fig:network}). A fully connected layer resulted in a 512 dimensional representation space. 

For each dataset, we set a non-informative hyperparameter $\lambda$ as the ratio between the number of images ($ |\mathcal{I}|$) and the number of image pairs ($|\mathcal{S}|$) to balance the number of samples in the two loss terms (Eq. \ref{eq:objective_function}). \rev{This ratio resulted in a reasonable balance between the two loss components in the objective function (see Supplement Fig. 3). }

\rev{We trained the models for 100 epochs using an Adam optimizer with an initial learning rate of 0.001. In the MR experiments, we reduced the learning rate by a factor of 0.2 if the training loss stopped decreasing for 5 epochs. We confirmed convergence based on the train loss curve in all experiments. The models were implemented in Keras 2.2.4 and ran on an Nvidia Quadro P6000 GPU with 2GB memory. Each training run in the MR experiments took approximately 2-4 hours.}

\textbf{Evaluation on the MR datasets.}  
For each dataset, we first trained LSSL \rev{on the collection of image pairs (subjects with only one MR were omitted from training) and applied the trained model to derive the brain age $\psi$ (projections along $\boldsymbol{\tau}$) for all the MRIs in the dataset. Note, the ground-truth diagnosis labels and the chronological age were omitted from the self-supervised training.} Then, the estimated brain age was correlated with chronological age in the control subjects to verify the quality of disentanglement. The brain age of control subjects was compared to that of the \rev{diagnosed} patients to reveal disease effects. Note, this type of analysis was unique to LSSL and could not be achieved by existing \rev{disentanglement approaches based on statistical formulation (e.g., \cite{kingma2013autoencoding,Higgins2017betaVAELB,oord2018representation,simclr}) because those methods do not estimate latent directions associated with underlying factors.}

LSSL also learned a representation vector for each MRI. We evaluated the quality of these representations by using them to classify diagnosis labels of individuals in a supervised setting. The classification was evaluated by 5-fold cross-validation, where the folds were split based on subjects; i.e., images of a single subject belonged to the same fold. After splitting folds, we performed the classification using both cross-sectional (i.e., based on single time points) and longitudinal models. The cross-sectional model discarded the temporal information within each subject and treated each image as an independent sample. The classifier was designed as a Multi-Layer Perceptron containing two fully connected layers of dimension 512 and 64 with ReLU activation. The longitudinal model was a Recurrent Neural Network (RNN), whose input was the longitudinal sequence of representation vectors for each subject. The RNN mapped each representation vector within the input sequence to a 16 dimensional vector, which was fed into a single layer GRU network with 16 hidden units \citep{ouyang2020recurrent} to predict the diagnosis label at each visit. In a separate experiment, we fine-tuned the LSSL representation by incorporating the encoder $\g$ (pre-trained by LSSL) into the classification models and then cross-validated this end-to-end classifier. \rev{Classification accuracy was measured by balanced accuracy (bAcc) to account for different number of samples in each cohort \citep{zhaoadeli2020cf-net}.}

\textbf{Baselines.} We compared the classification accuracy and speed of convergence of the end-to-end classifier to those whose encoders were pre-trained by several other state-of-the-art representation learning methods. For fair comparison, we used the same encoder architecture for all methods, as our goal was to show the superiority of our self-supervised representation learning rather than obtaining state-of-the-art results on any of the two dataset with more complex encoder architectures. As LSSL was conceptually related to a wide range of works, we selected several representative methods from unsupervised training (AE and VAE), factor disentanglement ($\beta$-VAE \citep{Higgins2017betaVAELB}), self-supervised learning (SimCLR \citep{simclr}), to a longitudinal framework based on Contrastive Predictive Coding (CPC \citep{oord2018representation}). Note, the pre-training of CPC already contained an auto-regressive model on top of the encoder, which reduced the representation to `context' features, i.e.,  a 16-D vector followed by a GRU with 16 hidden units. Therefore, the longitudinal prediction of CPC was directly based on the context features instead of the 512-D representation.

\subsection{Results of Synthetic Experiments}
\begin{figure}[!t]
    \centering
    
    \subfloat[]{\includegraphics[trim=0 -20 0 0, clip, width=0.47\linewidth]{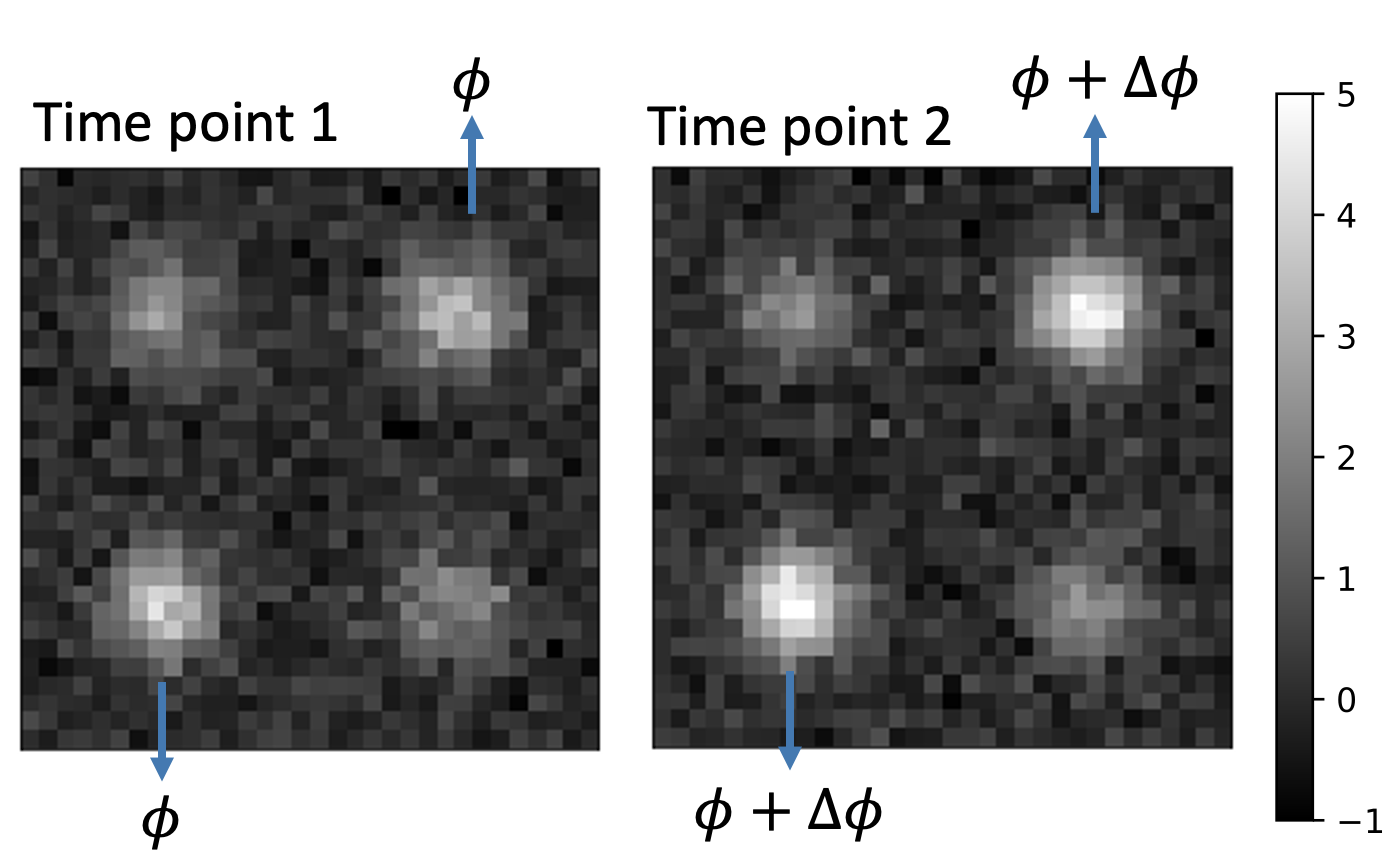} }
    \subfloat[]{\includegraphics[width=0.4\linewidth]{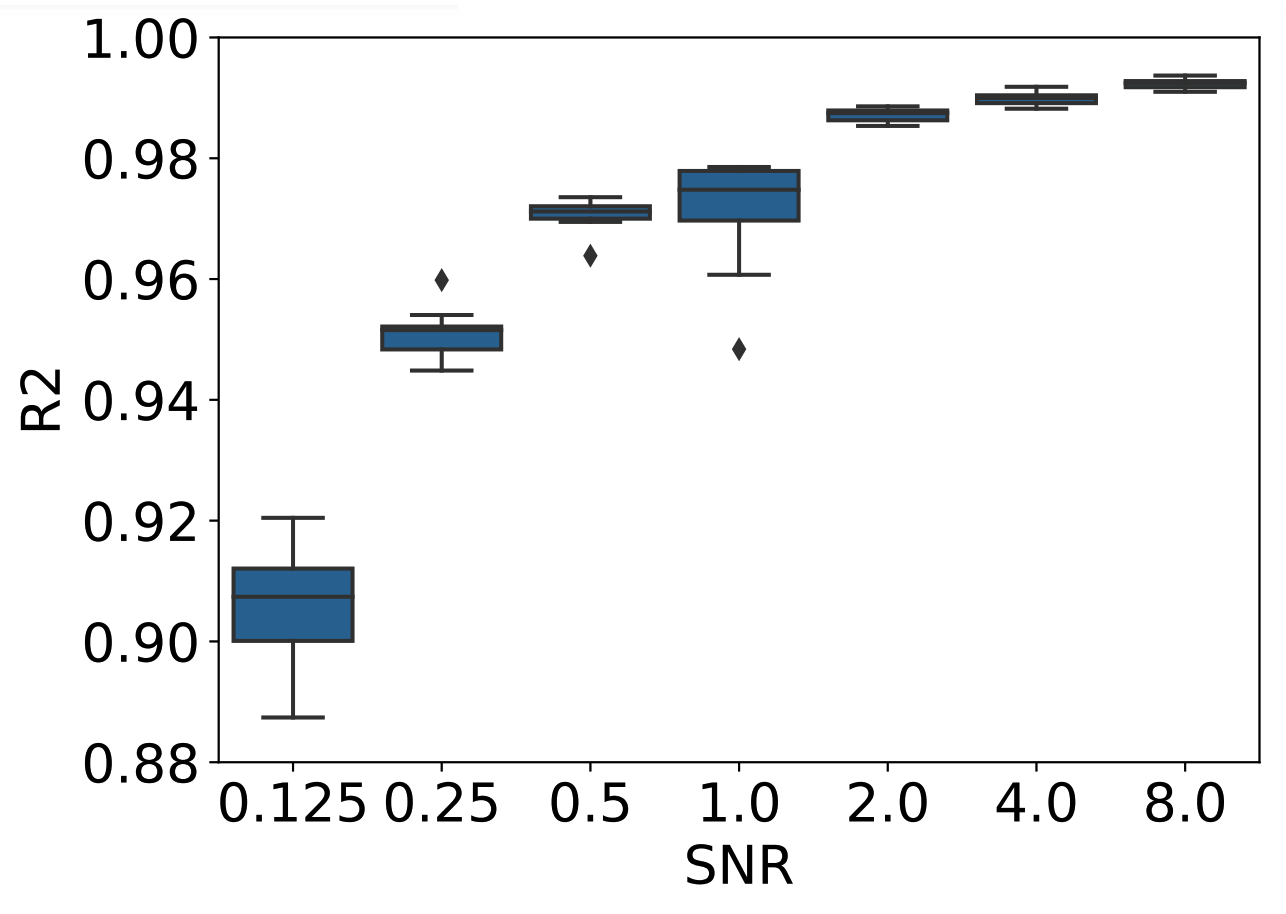}}\\
    \subfloat[]{\includegraphics[width=0.3\linewidth]{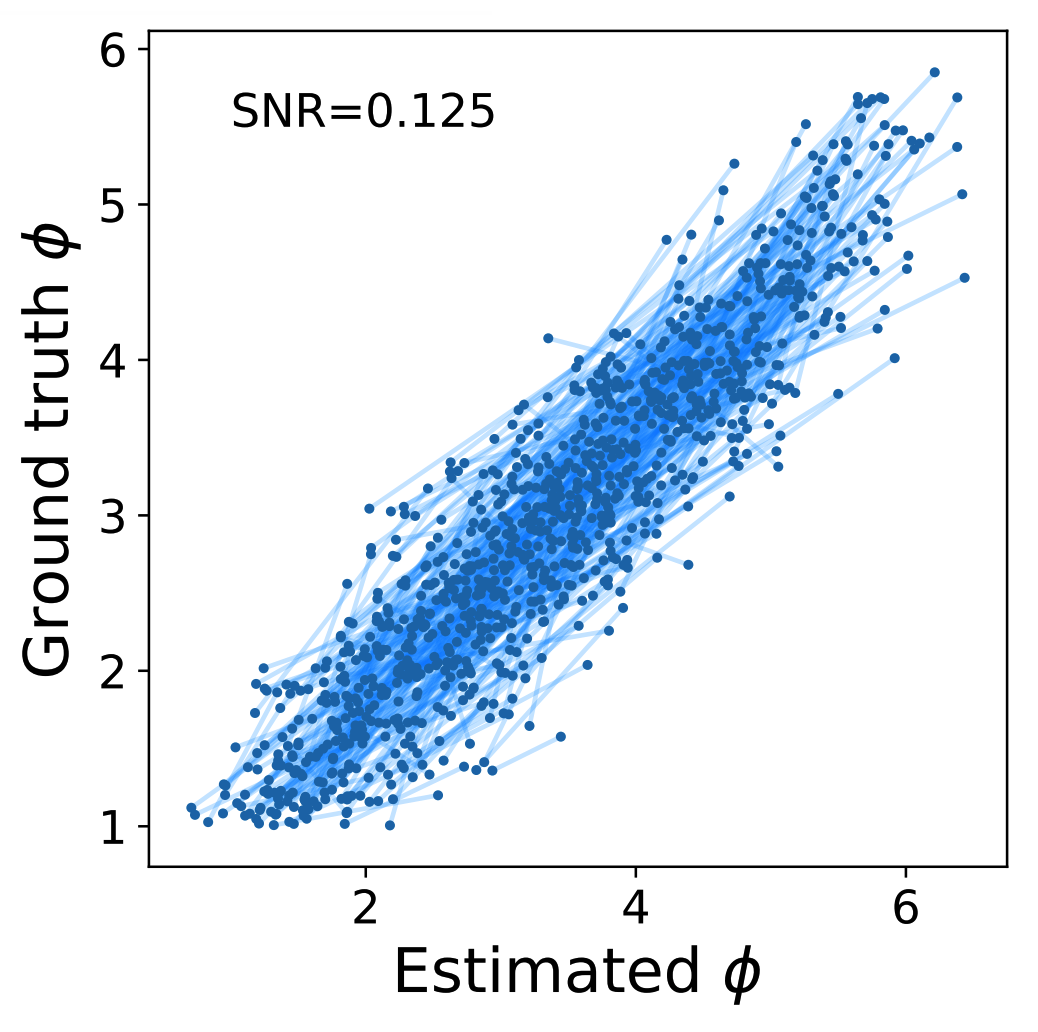}}
    \hspace{20pt}
    \subfloat[]{\includegraphics[trim=0 0 0 0, clip, width=0.35\linewidth]{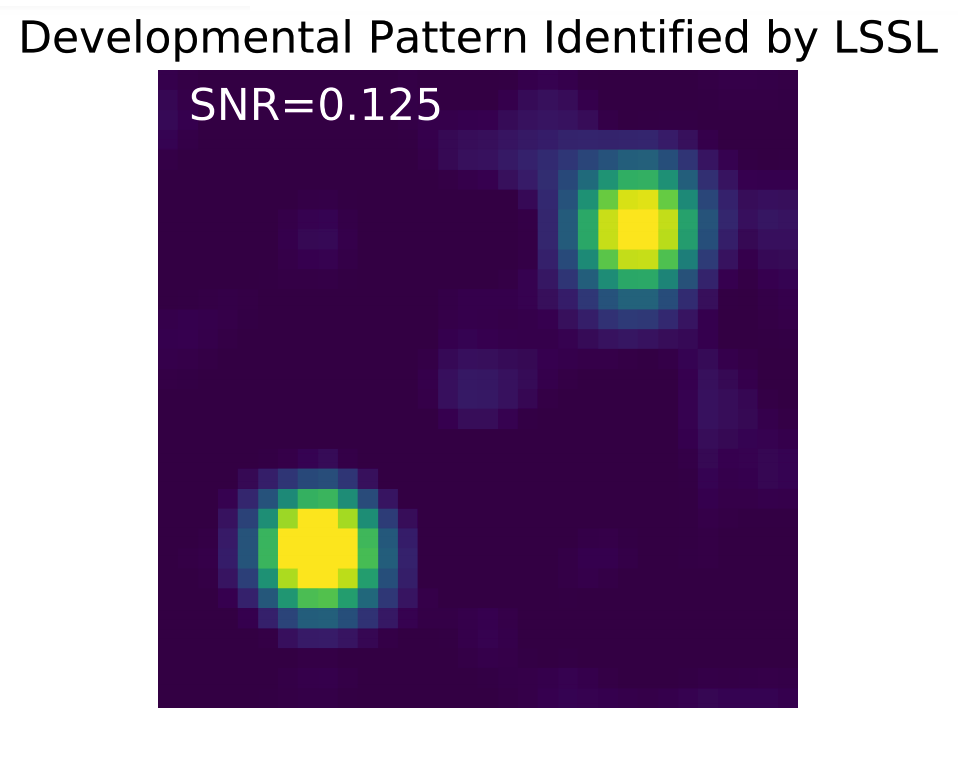}}
    \caption{\rev{(a) A synthetic image pair representing a subject's developmental effects, which were quantified by the magnitude change $\Delta \phi$ of the two off-diagonal Gaussians; (b) R2 between estimated and ground-truth $\phi$ with respect to noise level; (c) Group-level correlation for SNR=0.125. Each line connects two points associated with an image pair; (d) Identified developmental patterns for SNR=0.125.}}
    \label{fig:synthetic_results}
\end{figure}

\rev{We trained LSSL on the 512 image pairs and derived the brain age $\psi$ for each synthetic image.} As the scale of the coordinates in the latent space is not uniquely determined (one can rescale the latent space by rescaling the network parameters of the encoder and decoder), we normalized $\psi$ such that its mean and standard deviation were matched to \rev{those of the ground truth.} Note, this normalization was solely for intuitive interpretation of the results. \rev{Next, we quantified the correlation between the ground-truth and estimated $\psi$ by the R2 score \citep{Lewis05}. Fig. \ref{fig:synthetic_results}b indicates that this correlation was nearly perfect (R2=0.99) for a high SNR=8 and remained high for a low SNR=0.125. This global correlation in the range of $\psi \in [1,6]$ was learned from pairs of data whose maximum difference in brain age $\Delta\psi \leq 2$ (Fig. \ref{fig:synthetic_results}c). To identify the spatial pattern associated with increasing brain age, we varied the average latent representation $\bar{\boldsymbol{z}}$  (across all images) along $\boldsymbol{\tau}$ and $-\boldsymbol{\tau}$ by one unit and visualized the resulting difference between  the two reconstructed images:
\begin{equation}
    \textbf{d}(\bar{\boldsymbol{z}}+\boldsymbol{\tau};\phi) - \textbf{d}(\bar{\boldsymbol{z}}-\boldsymbol{\tau};\phi)
\end{equation}
The resulting image (Fig. \ref{fig:synthetic_results}d) shows that LSSL accurately estimated the developmental pattern in the two off-diagonal Gaussians even for the low SNR setting.}

\subsection{Longitudinal Study of Alzheimer’s Disease}
\begin{figure}[!t]
    \centering
    \begin{tikzpicture}
        \draw (0, 0) node[inner sep=0] {
    \includegraphics[width=0.47\linewidth]{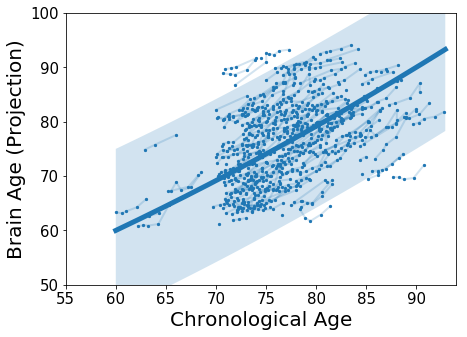} 
    \includegraphics[width=0.47\linewidth]{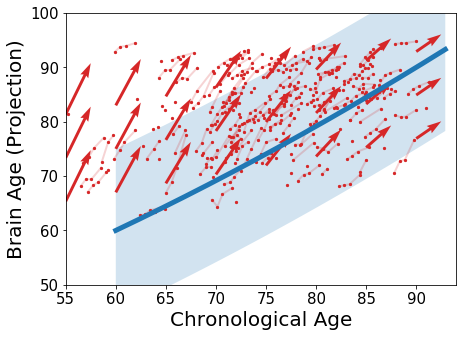}};
    \draw (-4.6, 1.55) node {\scriptsize \textbf{(a)}};
    \draw (1.25, 1.55) node {\scriptsize \textbf{(b)}};
    \end{tikzpicture}
    \begin{tikzpicture}
        \draw (0, 0) node[inner sep=0] {
    \includegraphics[width=0.24\linewidth]{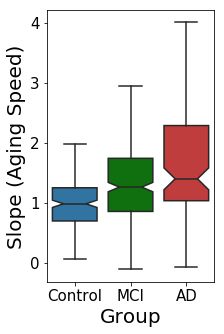}
    \includegraphics[width=0.36\linewidth]{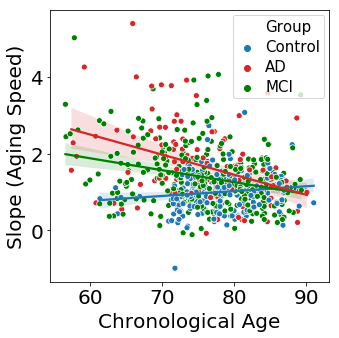}
    };
    \draw (-2.75, 1.75) node {\scriptsize \textbf{(c)}};
    \draw (0.3, 1.75) node {\scriptsize \textbf{(d)}};
    \end{tikzpicture}
    \caption{ \rev{(a) Brain age of 229 control subjects from ADNI1; (b) Brain age of 185 AD patients (red) overlaid with normal developmental trajectory (blue); (c) Speed of brain aging (slope of $\psi$ over time) for the 3 diagnosis groups and (d) as a function of chronological age of the ADNI1 subjects.} }
    \label{fig:adni_results}
\end{figure}

We trained LSSL on \rev{the $3,141$ pairs of MRIs from} ADNI and then applied the model to derive the brain age $\psi$ for all 2,641 MRI in the dataset. \rev{The mean and standard deviation of the estimated $\psi$ were normalized according to the} chronological age in the dataset. Fig. \ref{fig:adni_results}a shows the brain age of the control subjects versus their chronological age. According to the fitting of a quadratic mixed effect model, brain age and chronological age exhibited a nearly linear relationship over the entire age span of the dataset. \rev{To ensure this correlation was not a result of model overfitting, we measured the Pearson's correlation only on the 43 control MRIs that were not a part of the training set (see Supplement Fig. 4).} The `global' correlation \rev{in the range of 60 to 90 years} was derived solely from the ordinal information from subject-specific image pairs (maximum 4 years part) without using the ground-truth chronological age of subjects. 

\begin{figure*}[!t]
    \centering
    \includegraphics[width=\linewidth]{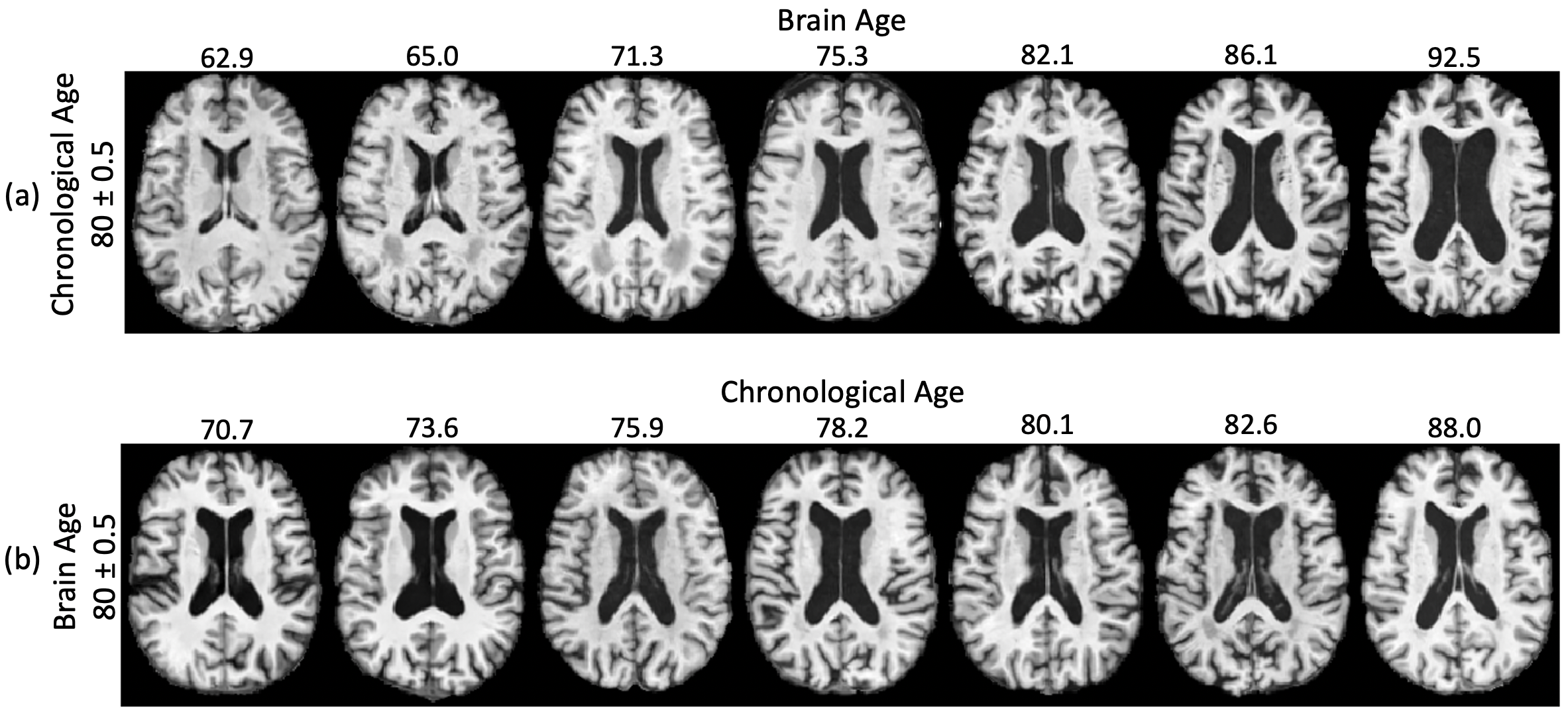} 
    \caption{(a) Control subjects with the same chronological age of 80 years had distinct brain appearance and brain age estimated by LSSL. (b) Control subjects with the same brain age of 80 years exhibited similar brain appearance. }
    \label{fig:age_vs_brainage}
\end{figure*}

Next, we qualitatively show that the estimated $\psi$ was a more accurate marker for brain age compared to the chronological age. Fig. \ref{fig:age_vs_brainage}a displays the brain of 7 normal control subjects with the same chronological age (of 80 years), yet their estimated brain age ranged from 62.9 years to 92.5 years. \rev{This large variance visually relates to the difference in ventricle size and cortical thickness shown in Fig. \ref{fig:age_vs_brainage}a. It is also consistent with the variance of brain age plotted in Fig. \ref{fig:adni_results}a for any given chronological age. On the other hand, control subjects with the same estimated brain age shown in Fig. \ref{fig:age_vs_brainage}b had} similar brain appearance despite that they had various chronological age.  These results not only support the efficacy of our brain age estimation but also suggest that supervised training based on chronological age may be a flawed strategy for learning the brain age even within the control cohort.

According to Fig. \ref{fig:adni_results}b, the brain age of AD patients was generally higher than chronological age reflecting the neurodegenerative nature of AD that accelerated brain aging. This phenomenon can also be inferred from Fig. \ref{fig:adni_results}c, where we computed \rev{the `brain age slope'} by fitting a simple linear regression for each subject (with at least 2 images) on their brain age across visits. In doing so, we see that the control group had an average aging speed (slope) close to 1, indicating the consistency between the progression rate of brain age and of chronological age. In comparison, brain aging of the AD group was significantly faster ($p<0.001$, two-sample $t$-test). Interestingly, the MCI group, representing a transitional state between control and AD, had an intermediate aging speed, which was significantly faster than normal and slower than AD ($p<0.001$, two-sample $t$-tests). Moreover, we observe that the gap between brain age and chronological age was larger in younger AD patients than the older ones (Fig. \ref{fig:adni_results}b), which indicates the disease effect was more prominent in the younger brain. This phenomenon was quantitatively supported by fitting a linear regression between the aging speed (slope) and age in each cohort (Fig. \ref{fig:adni_results}d). While the AD subjects had more accelerated aging at younger ages, their aging speed was not different from that of the oldest old in the control cohort. Again, the MCI group exhibited an intermediate effect between AD and control subjects. \rev{Lastly, to ensure that these results were not specific to the resolution of input images, we repeated the above experiments based on the $80 \times 80 \times 80$ input resolution, which resulted in similar findings (Supplement Fig. 1).}

\begin{figure*}[!t]
    \centering
    \includegraphics[width=\linewidth]{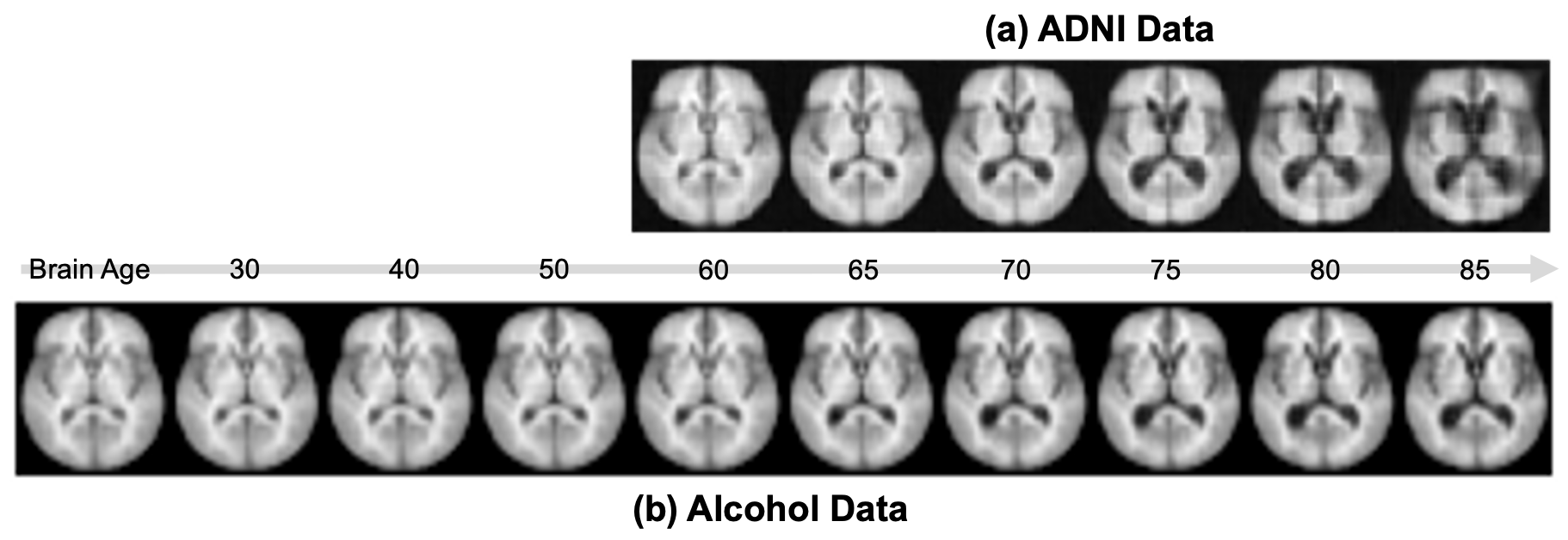} 
    \caption{Simulated average brain at different brain ages from the ADNI and alcohol data. }
    \label{fig:adni_aud_brain}
\end{figure*}

We also assessed the quality of disentanglement by simulating the average brain at different brain ages. We constructed the average representation associated brain age $\psi$ via 
\begin{equation}
\psi\boldsymbol{\tau} + \frac{1}{N}\sum_{i=1}^N \left(\z^i - \z^i\boldsymbol{\tau}^\top\boldsymbol{\tau}\right),
\end{equation}
where the first term corresponds to the representation of brain age and the second term captures factors independent from brain age, i.e., the group average of the components orthogonal to $\boldsymbol{\tau}$. By decoding this age-dependent representation, we observe a pattern of enlargement in the ventricle and loss of brain tissues as age increases (Fig. \ref{fig:adni_aud_brain}). This pattern converges with the current understanding of brain aging in the neuroscience literature \citep{Sullivan08}.

\begin{table}[!t]
  \caption{ADNI Data set: \rev{Balanced accuracy (bAcc)} of cross-sectional and longitudinal classification with and w/o fine-tuning the encoder. Best result in each column is in bold and the second best is underlined.}
  \label{table:adni}
  \centering
  \begin{tabular}{l c c }
    \toprule
    & \multicolumn{2}{c}{ADNI}  \\
    \cline{2-3} 
    Pre-training & CNN & CNN+RNN \\
    \cline{2-3} 
    Model     & frozen / fine-tuned & frozen / fine-tuned \\
    \hline
    No pretrain & - / 80.6 & - / 74.6 \\
    AE & 58.6 / 81.7 & 62.1 / 71.3 \\
    VAE \citep{kingma2013autoencoding} & 58.9 / 75.7 & 62.8 / 71.9 \\
    $\beta$-VAE \citep{Higgins2017betaVAELB} & 56.1 / 77.2 & 76.3 / 78.4 \\
    SimCLR \citep{simclr} & \textbf{78.0} / \textbf{84.4} & \underline{80.7} / \underline{84.7} \\
    CPC \citep{oord2018representation} & 65.5 / 78.6 & 66.7 / 80.4  \\
    Ours (LSSL) & \underline{72.0} / \underline{84.1} & \textbf{81.8} / \textbf{87.0}  \\
    \bottomrule
  \end{tabular}
\end{table}
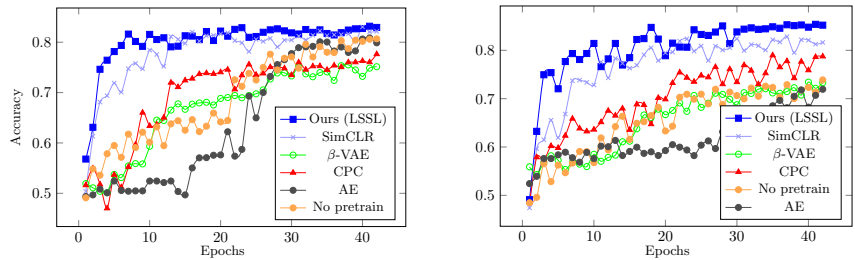
\begin{figure}[!t]
    \centering
\begin{tabular}{cc}
    \resizebox {0.45\textwidth} {!} {
\begin{tikzpicture}
    \begin{axis}[
	width=0.8\linewidth,
	height=0.55\linewidth,
        xlabel= {\small Epochs},
        ylabel= {\small Accuracy},
        xlabel style={yshift=0.2cm,xshift=-0.2cm},
        ylabel style={yshift=-0.3cm},
	legend pos =  south east
]
    \addplot[smooth,color=blue,mark=square*,error bars/.cd,y dir=both,y explicit]
        plot coordinates {
(1,0.568020)
(2,0.631140)
(3,0.746300)
(4,0.764340)
(5,0.781460)
(6,0.793620)
(7,0.816180)
(8,0.801560)
(9,0.791820)
(10,0.815540)
(11,0.805340)
(12,0.808820)
(13,0.790800)
(14,0.792200)
(15,0.812800)
(16,0.812200)
(17,0.805200)
(18,0.821200)
(19,0.803400)
(20,0.822200)
(21,0.812000)
(22,0.825600)
(23,0.828800)
(24,0.809800)
(25,0.811400)
(26,0.819800)
(27,0.824600)
(28,0.826600)
(29,0.822600)
(30,0.818600)
(31,0.816400)
(32,0.825200)
(33,0.819600)
(34,0.824600)
(35,0.817400)
(36,0.814400)
(37,0.819000)
(38,0.824200)
(39,0.826600)
(40,0.827200)
(41,0.831800)
(42,0.829400)
        };
    \addlegendentry{\small{Ours (LSSL)}}
    
    \addplot[smooth,color=blue!40,mark=x,error bars/.cd,y dir=both,y explicit,]
        plot coordinates {
(1,0.503620)
(2,0.613200)
(3,0.681100)
(4,0.693780)
(5,0.718860)
(6,0.700880)
(7,0.747436)
(8,0.757920)
(9,0.747280)
(10,0.784500)
(11,0.775660)
(12,0.752242)
(13,0.811210)
(14,0.805780)
(15,0.797740)
(16,0.820180)
(17,0.801900)
(18,0.812880)
(19,0.813546)
(20,0.816480)
(21,0.819020)
(22,0.803220)
(23,0.798900)
(24,0.781442)
(25,0.805520)
(26,0.801620)
(27,0.807620)
(28,0.790780)
(29,0.803960)
(30,0.804400)
(31,0.811960)
(32,0.804700)
(33,0.813754)
(34,0.822620)
(35,0.813060)
(36,0.814760)
(37,0.815100)
(38,0.816480)
(39,0.802100)
(40,0.830520)
(41,0.820600)
(42,0.821560)
        };
    \addlegendentry{\small{SimCLR}}
    
    \addplot[smooth,color=green,mark=o,error bars/.cd,y dir=both,y explicit,]
        plot coordinates {
(1,0.519260)
(2,0.511000)
(3,0.504000)
(4,0.501200)
(5,0.537400)
(6,0.531200)
(7,0.554400)
(8,0.559000)
(9,0.558400)
(10,0.593400)
(11,0.645400)
(12,0.644600)
(13,0.665200)
(14,0.677600)
(15,0.666800)
(16,0.677000)
(17,0.679400)
(18,0.680800)
(19,0.675400)
(20,0.688400)
(21,0.690200)
(22,0.694200)
(23,0.689800)
(24,0.694200)
(25,0.697600)
(26,0.705400)
(27,0.727000)
(28,0.738800)
(29,0.739200)
(30,0.733000)
(31,0.751200)
(32,0.737200)
(33,0.731400)
(34,0.739800)
(35,0.741000)
(36,0.724200)
(37,0.753400)
(38,0.755000)
(39,0.745000)
(40,0.732200)
(41,0.748600)
(42,0.751400)

        };
    \addlegendentry{\small{$\beta$-VAE}}
    
    \addplot[smooth,color=red,mark=triangle*,error bars/.cd,y dir=both,y explicit,]
        plot coordinates {
(1,0.516180)
(2,0.549400)
(3,0.517760)
(4,0.470160)
(5,0.536800)
(6,0.512000)
(7,0.552400)
(8,0.596600)
(9,0.660200)
(10,0.633800)
(11,0.636400)
(12,0.650400)
(13,0.719800)
(14,0.711000)
(15,0.724200)
(16,0.727800)
(17,0.736800)
(18,0.738600)
(19,0.737200)
(20,0.740400)
(21,0.745600)
(22,0.706400)
(23,0.734400)
(24,0.755800)
(25,0.735400)
(26,0.738000)
(27,0.750600)
(28,0.750200)
(29,0.748200)
(30,0.734800)
(31,0.760000)
(32,0.745600)
(33,0.752280)
(34,0.752600)
(35,0.745200)
(36,0.753800)
(37,0.750600)
(38,0.760200)
(39,0.760200)
(40,0.762800)
(41,0.760400)
(42,0.776380)

        };
    \addlegendentry{\small{CPC}}
        
    \addplot[smooth,black!70,mark=*,  error bars/.cd,
    y dir=both,
    y explicit,
] plot coordinates {
(1,0.493540)
(2,0.496800)
(3,0.508400)
(4,0.501000)
(5,0.524400)
(6,0.505600)
(7,0.504200)
(8,0.504800)
(9,0.504800)
(10,0.524400)
(11,0.524400)
(12,0.521400)
(13,0.524400)
(14,0.503600)
(15,0.496840)
(16,0.550400)
(17,0.570400)
(18,0.570800)
(19,0.575600)
(20,0.576200)
(21,0.622200)
(22,0.573800)
(23,0.587200)
(24,0.693000)
(25,0.649800)
(26,0.700200)
(27,0.732000)
(28,0.755600)
(29,0.765200)
(30,0.777400)
(31,0.789200)
(32,0.790800)
(33,0.791800)
(34,0.800800)
(35,0.800400)
(36,0.782600)
(37,0.790440)
(38,0.779700)
(39,0.782700)
(40,0.806400)
(41,0.808580)
(42,0.798800)

    };
    \addlegendentry{\small{AE}}

    \addplot[smooth,orange!70,mark=*,  error bars/.cd,
    y dir=both,
    y explicit,
] plot coordinates {
(1,0.491000)
(2,0.548360)
(3,0.535900)
(4,0.578500)
(5,0.594900)
(6,0.571480)
(7,0.617496)
(8,0.590900)
(9,0.621400)
(10,0.601480)
(11,0.631880)
(12,0.594514)
(13,0.637890)
(14,0.644080)
(15,0.625280)
(16,0.646280)
(17,0.622480)
(18,0.633280)
(19,0.659286)
(20,0.641700)
(21,0.644680)
(22,0.725400)
(23,0.712480)
(24,0.738400)
(25,0.724900)
(26,0.750080)
(27,0.776260)
(28,0.744920)
(29,0.768280)
(30,0.770880)
(31,0.748400)
(32,0.795800)
(33,0.770680)
(34,0.791260)
(35,0.780060)
(36,0.779460)
(37,0.803080)
(38,0.787380)
(39,0.804100)
(40,0.801680)
(41,0.805860)
(42,0.806800)

    };
    \addlegendentry{\small{No pretrain}}
    
    \end{axis}
    \end{tikzpicture}
    } & 
    \resizebox {0.43\textwidth} {!} {
\begin{tikzpicture}
    \begin{axis}[
	width=0.8\linewidth,
	height=0.55\linewidth,
        xlabel= {\small Epochs},
        xlabel style={yshift=0.2cm,xshift=-0.2cm},
	legend pos =  south east
]
    \addplot[smooth,color=blue,mark=square*,error bars/.cd,y dir=both,y explicit]
        plot coordinates {
(1,0.49104	)
(2,0.6324	)
(3,0.7494	)
(4,0.7538	)
(5,0.7204	)
(6,0.7768	)
(7,0.7932	)
(8,0.781 	)
(9,0.7934	)
(10,0.8142	)
(11,0.7662	)
(12,0.7822	)
(13,0.8142	)
(14,0.7696	)
(15,0.7848	)
(16,0.8224	)
(17,0.8242	)
(18,0.8474	)
(19,0.8232	)
(20,0.7888	)
(21,0.8148	)
(22,0.8068	)
(23,0.8064	)
(24,0.8424	)
(25,0.8328	)
(26,0.8308	)
(27,0.8378	)
(28,0.8504	)
(29,0.8138	)
(30,0.838 	)
(31,0.8438	)
(32,0.8464	)
(33,0.8434	)
(34,0.8444	)
(35,0.8486	)
(36,0.8457	)
(37,0.8528	)
(38,0.851 	)
(39,0.8524	)
(40,0.84788	)
(41,0.85374	)
(42,0.85206	)
        };
    \addlegendentry{\small{Ours (LSSL)}}
    \addplot[smooth,color=blue!40,mark=x,error bars/.cd,y dir=both,y explicit,]
        plot coordinates {
(1, 0.474)
(2, 0.5964)
(3, 0.6152)
(4, 0.6554)
(5, 0.6194)
(6, 0.6682)
(7, 0.737)
(8, 0.7384)
(9, 0.7338)
(10,0.7284)
(11,0.7758)
(12,0.743)
(13,0.783)
(14,0.773)
(15,0.763)
(16,0.7692)
(17,0.7906)
(18,0.8052)
(19,0.7788)
(20,0.7948)
(21,0.7964)
(22,0.819)
(23,0.8238)
(24,0.7984)
(25,0.8088)
(26,0.7788)
(27,0.8086)
(28,0.8006)
(29,0.8054)
(30,0.8238)
(31,0.7996)
(32,0.8118)
(33,0.8096)
(34,0.8006)
(35,0.8218)
(36,0.817)
(37,0.8282)
(38,0.7952)
(39,0.8218)
(40,0.8208	)
(41,0.8126	)
(42,0.8162	)
        };
    \addlegendentry{\small{SimCLR}}
    \addplot[smooth,color=green,mark=o,error bars/.cd,y dir=both,y explicit,]
        plot coordinates {
(1, 0.55902)
(2, 0.54302)
(3, 0.56782)
(4, 0.58182)
(5, 0.57262)
(6, 0.55342)
(7, 0.56842)
(8, 0.56462)
(9, 0.55942)
(10,0.58482)
(11,0.57042)
(12,0.57848)
(13,0.58322)
(14,0.61082)
(15,0.61202)
(16,0.63762)
(17,0.66622)
(18,0.66082)
(19,0.67822)
(20,0.66682)
(21,0.67562)
(22,0.69102)
(23,0.67382)
(24,0.70662)
(25,0.68202)
(26,0.69062)
(27,0.71122)
(28,0.70882)
(29,0.69862)
(30,0.68742)
(31,0.71242)
(32,0.71082)
(33,0.71962)
(34,0.72062)
(35,0.71182)
(36,0.71102)
(37,0.71462)
(38,0.72262)
(39,0.70742)
(40,0.73422	)
(41,0.72142	)
(42,0.73302	)
        };
    \addlegendentry{\small{$\beta$-VAE}}
    \addplot[smooth,color=red,mark=triangle*,error bars/.cd,y dir=both,y explicit,]
        plot coordinates {
(1, 0.48546)
(2, 0.5788)
(3, 0.5728)
(4, 0.60204)
(5, 0.59802)
(6, 0.6234)
(7, 0.6586)
(8, 0.6384)
(9, 0.63228)
(10,0.636)
(11,0.6514)
(12,0.675)
(13,0.665)
(14,0.6796)
(15,0.63558)
(16,0.6882)
(17,0.68692)
(18,0.64726)
(19,0.7016)
(20,0.6986)
(21,0.7324)
(22,0.75414)
(23,0.7434)
(24,0.735)
(25,0.7478)
(26,0.7442)
(27,0.7654)
(28,0.73)
(29,0.761)
(30,0.7342)
(31,0.7692)
(32,0.758)
(33,0.73)
(34,0.7526)
(35,0.7896)
(36,0.765)
(37,0.7776)
(38,0.7424)
(39,0.7766)
(40,0.7588)
(41,0.7862)
(42,0.78738	)
        };
    \addlegendentry{\small{CPC}}
    \addplot[smooth,orange!70,mark=*,  error bars/.cd,
    y dir=both,
    y explicit,
] plot coordinates {
(1, 0.4847)
(2, 0.49596)
(3, 0.56536)
(4, 0.52836)
(5, 0.56256)
(6, 0.54676)
(7, 0.56676)
(8, 0.59076)
(9, 0.59036)
(10,0.56756)
(11,0.61876)
(12,0.59396)
(13,0.64218)
(14,0.66354)
(15,0.61574)
(16,0.64914)
(17,0.65234)
(18,0.66494)
(19,0.68214)
(20,0.63314)
(21,0.64314)
(22,0.70274)
(23,0.70914)
(24,0.69894)
(25,0.70934)
(26,0.68914)
(27,0.71954)
(28,0.68774)
(29,0.7135)
(30,0.70634)
(31,0.69954)
(32,0.72194)
(33,0.70494)
(34,0.72574)
(35,0.72334)
(36,0.70294)
(37,0.72894)
(38,0.70014)
(39,0.71794)
(40,0.72334)
(41,0.71354)
(42,0.73874)
    };
    \addlegendentry{\small{No pretrain}}
    \addplot[smooth,black!70,mark=*,  error bars/.cd,
    y dir=both,
    y explicit,
] plot coordinates {
(1, 0.52428)
(2, 0.5394)
(3, 0.57614)
(4, 0.57604)
(5, 0.58382)
(6, 0.58922)
(7, 0.5778)
(8, 0.5686)
(9, 0.5888)
(10,0.576)
(11,0.6018)
(12,0.6006)
(13,0.6136)
(14,0.5824)
(15,0.5908)
(16,0.5996)
(17,0.5868)
(18,0.59)
(19,0.5824)
(20,0.5926)
(21,0.6048)
(22,0.5996)
(23,0.5938)
(24,0.5822)
(25,0.6038)
(26,0.6126)
(27,0.5968)
(28,0.6314)
(29,0.5966)
(30,0.651)
(31,0.634)
(32,0.6492)
(33,0.6594)
(34,0.6744)
(35,0.6852)
(36,0.6964)
(37,0.6812)
(38,0.7048)
(39,0.71794)
(40,0.68244)
(41,0.70696)
(42,0.7194)
    };
    \addlegendentry{\small{AE}}
    \end{axis}
    \end{tikzpicture}
    } \\
    {\footnotesize (a) Single time point classification (CNN)} &
    {\footnotesize (b) Longitudinal classification  (CNN+RNN)} \\ 
\end{tabular}
    \caption{ADNI Dataset - Average bAcc in the first 45 epochs over the 5 testing folds when training end-to-end classification based on pretrained encoders.}
    \label{fig:adni_convergence}
\end{figure}

Lastly, we classified AD patients from control subjects based on the learned representations. In the cross-sectional setting, the representations learned by LSSL enabled more accurate prediction than the baselines except for SimCLR (Table \ref{table:adni} CNN), and the \rev{bAcc} of our model closely matched up to SimCLR after fine-tuning the encoder in the cross-sectional setting. On the other hand, when performing classification based on the longitudinal sequences (CNN+RNN), the \rev{bAcc} associated with the LSSL representations increased and outperformed all baselines, which was also the case after fine-tuning the encoders. \rev{This accuracy improvement over cross-sectional CNN indicates that LSSL resulted in informative temporal trajectories of individuals in the representation space, which could only be learned by the RNN models.}
Moreover, when performing the fine-tuning in an end-to-end setting, the encoder pre-trained by LSSL converged the fastest in both cross-sectional and longitudinal settings (see Fig. \ref{fig:adni_convergence}) despite that all methods started from random predictions (as the MLP layers were randomly initialized). \rev{When using the most accurate implementation (LSSL+RNN+fine-tuning), the bAcc for distinguishing MCI from controls (69.9\%) and  AD (69.5\%)  were also higher than those without pretraining of LSSL (68.3\% and 66.5\%). Note, the classification accuracies are in line with the  literature \citep{Oh2019}, which ranges from 60\% to 75\% bAcc depending on the progressiveness of the MCI subjects used for classification.}

\subsection{Longitudinal Study of Alcohol Dependence}
\begin{figure}[!t]
    \centering
    \begin{tikzpicture}
        \draw (0, 0) node[inner sep=0] {
    \includegraphics[width=0.47\linewidth]{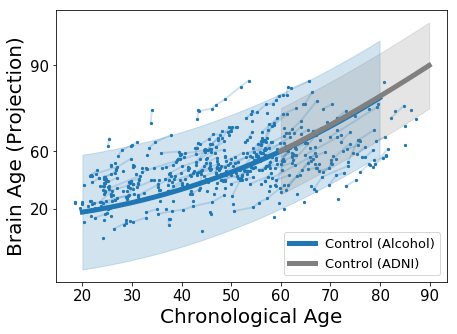} 
    \includegraphics[width=0.47\linewidth]{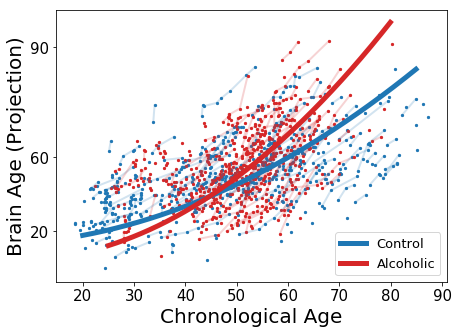}};
    \draw (-4.6, 1.55) node {\scriptsize \textbf{(a)}};
    \draw (1.25, 1.55) node {\scriptsize \textbf{(b)}};
    \end{tikzpicture}
    \begin{tikzpicture}
        \draw (0, 0) node[inner sep=0] {
    \includegraphics[width=0.26\linewidth]{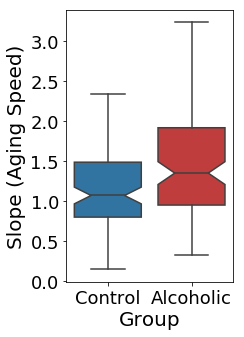}
    \includegraphics[width=0.36\linewidth]{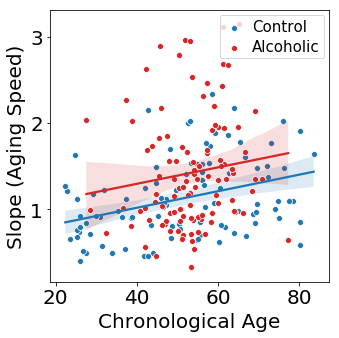}};
    \draw (-2.7, 1.75) node {\scriptsize \textbf{(c)}};
    \draw (0.35, 1.75) node {\scriptsize \textbf{(d)}};
    \end{tikzpicture}
    \caption{(a) Brain age of 274 normal control subjects from the alcohol dataset (blue) overlaid with the ADNI1 controls (gray); (b) Brain age of 329 alcohol-dependent subjects (red); (c,d) Speed of brain aging for all subjects with two or more images in the alcohol data set. }
    \label{fig:aud_results}
\end{figure}

Similar to the previous experiment, we first trained LSSL to estimate the brain age for all MR images and then visualized the brain age of the control subjects (Fig. \ref{fig:aud_results}a). To put the results in line with the ADNI experiments, we normalized the projection $\psi$ with respect to the age range of the ADNI dataset \rev{(i.e., confined to 60 to 90 years and then applied to the entire age range of the alcohol data set, Fig. \ref{fig:aud_results}a)}. 
Compared to the approximately linear aging pattern between age 60 to 90, the aging of the control subjects exhibited a quadratic pattern \rev{ over a longer life span, where the aging speed was slower for younger subjects (e.g., between 20 and 60 years) compared to the older subjects (after 60 years, Fig. \ref{fig:aud_results}a)}. Similar to the ADNI experiment, \rev{the alcoholics} also exhibited higher brain age than normal controls (Fig. \ref{fig:aud_results}b), which was supported by their slopes (aging speed) being significant larger than normal ($p<0.001$, Fig. \ref{fig:aud_results}c). Different from the ADNI experiment was that the aging speed of \rev{alcohol-dependent subjects} was always faster than normal controls regardless of their chronological age (Fig. \ref{fig:aud_results}d). Furthermore, older subjects had a larger gap between brain and chronological age (Fig. \ref{fig:aud_results}b), which indicates an accumulative alcohol effect. The accumulative alcohol effect is also supported by the observation that chronic drinking gradually deteriorates brain structure resulting in more severe alcohol effect in older subjects \citep{Zhao19AB}.

In Fig. \ref{fig:adni_aud_brain},  we simulated images of different brain ages for the alcohol dataset. We observe that the simulated brains closely mimic the ones from the ADNI experiment from age 60 to 70, a range where the two datasets highly overlap. However, the simulated brains from the alcohol experiment showed less pronounced aging effect after age 70 compared to the ADNI results. This was potentially due to the few older subjects in the alcohol dataset compared with ADNI, so the model conservatively extrapolated the aging pattern for the older age range.

\begin{figure}[!t]
    \centering
\begin{tabular}{cc}
    \resizebox {0.45\textwidth} {!} {
\begin{tikzpicture}
    \begin{axis}[
	width=0.8\linewidth,
	height=0.55\linewidth,
        xlabel= {\small Epochs},
        ylabel={\small Accuracy},
        xlabel style={yshift=0.2cm,xshift=-0.2cm},
        ylabel style={yshift=-0.3cm},
	legend pos =  south east
]
    \addplot[smooth,color=blue,mark=square*,error bars/.cd,y dir=both,y explicit]
        plot coordinates {

(1,0.5)
(2,0.5)
(3,0.5058)
(4,0.5238)
(5,0.5174)
(6,0.5528)
(7,0.5522)
(8,0.5438)
(9,0.60494)
(10,0.5894)
(11,0.6089)
(12,0.6032)
(13,0.6076)
(14,0.5924)
(15,0.6198)
(16,0.6132)
(17,0.6302)
(18,0.6519)
(19,0.6384)
(20,0.6604)
(21,0.65262)
(22,0.6436)
(23,0.66)
(24,0.6626)
(25,0.6775)
(26,0.67296)
(27,0.67764)
(28,0.67864)
(29,0.66320)
(30,0.6882)
(31,0.66956)
(32,0.7058)
(33,0.69116)
(34,0.7026)
(35,0.67864)
(36,0.689340)
(37,0.687)
(38,0.6877)
(39,0.6982)
(40,0.67622)
(41,0.695)
(42,0.69720)
        };
    \addlegendentry{\small{Ours (LSSL)}}
    
    \addplot[smooth,color=blue!40,mark=x,error bars/.cd,y dir=both,y explicit,]
        plot coordinates {
(1,0.4996)
(2,0.4822)
(3,0.5036)
(4,0.5034)
(5,0.5256)
(6,0.50832)
(7,0.5476)
(8,0.57866)
(9,0.5754)
(10,0.5868)
(11,0.6066)
(12,0.595)
(13,0.58608)
(14,0.6002)
(15,0.6156)
(16,0.598560)
(17,0.63880)
(18,0.6260)
(19,0.62528)
(20,0.6376)
(21,0.6402)
(22,0.6548)
(23,0.6408)
(24,0.6518)
(25,0.64040)
(26,0.65596)
(27,0.64582)
(28,0.6676)
(29,0.647)
(30,0.658)
(31,0.6511)
(32,0.6378)
(33,0.64152)
(34,0.65396)
(35,0.636)
(36,0.6562)
(37,0.6238)
(38,0.6375)
(39,0.654)
(40,0.6236)
(41,0.65176)
(42,0.6318)
        };
    \addlegendentry{\small{SimCLR}}
    
    \addplot[smooth,color=green,mark=o,error bars/.cd,y dir=both,y explicit,]
        plot coordinates {
(1,0.5208)
(2,0.49684)
(3,0.48294)
(4,0.49762)
(5,0.5001)
(6,0.4938)
(7,0.50034)
(8,0.50402)
(9,0.50722)
(10,0.53118)
(11,0.509452)
(12,0.527454)
(13,0.53938)
(14,0.551136)
(15,0.58172)
(16,0.557725)
(17,0.58844)
(18,0.5707)
(19,0.58856)
(20,0.59184)
(21,0.59192)
(22,0.60084)
(23,0.61196)
(24,0.6037)
(25,0.625284)
(26,0.63026)
(27,0.6269)
(28,0.6258)
(29,0.64416)
(30,0.613709)
(31,0.635825)
(32,0.629770)
(33,0.648966)
(34,0.64322)
(35,0.65174)
(36,0.61751)
(37,0.651186)
(38,0.62862)
(39,0.6566)
(40,0.630994)
(41,0.6241935)
(42,0.64062)

        };
    \addlegendentry{\small{$\beta$-VAE}}
    
    \addplot[smooth,color=red,mark=triangle*,error bars/.cd,y dir=both,y explicit,]
        plot coordinates {
(1,0.49080)
(2,0.49540)
(3,0.49980)
(4,0.48072)
(5,0.4984)
(6,0.52112)
(7,0.5036)
(8,0.528844)
(9,0.536534)
(10,0.542712)
(11,0.53252)
(12,0.535045)
(13,0.54368)
(14,0.56120)
(15,0.56578)
(16,0.555394)
(17,0.59078)
(18,0.576622)
(19,0.5836)
(20,0.603862)
(21,0.605574)
(22,0.59726)
(23,0.61432)
(24,0.58888)
(25,0.6014)
(26,0.58742)
(27,0.6134)
(28,0.58814)
(29,0.625562)
(30,0.60398)
(31,0.62206)
(32,0.61296)
(33,0.619)
(34,0.62288)
(35,0.61562)
(36,0.632386)
(37,0.6474275)
(38,0.631134)
(39,0.642336)
(40,0.637502)
(41,0.638824)
(42,0.64226)

        };
    \addlegendentry{\small{CPC}}
        
    \addplot[smooth,black!70,mark=*,  error bars/.cd,
    y dir=both,
    y explicit,
] plot coordinates {
(1,0.4962)
(2,0.5)
(3,0.5092)
(4,0.53234)
(5,0.586)
(6,0.55536)
(7,0.604)
(8,0.6121)
(9,0.6092)
(10,0.624)
(11,0.6412)
(12,0.64282)
(13,0.6457)
(14,0.6424)
(15,0.641)
(16,0.6452)
(17,0.6404)
(18,0.63642)
(19,0.6428)
(20,0.6638)
(21,0.65512)
(22,0.64102)
(23,0.65896)
(24,0.648234)
(25,0.657406)
(26,0.65894)
(27,0.65812)
(28,0.65712)
(29,0.64196)
(30,0.6426)
(31,0.66086)
(32,0.647906)
(33,0.64644)
(34,0.6501)
(35,0.64806)

(36,0.648562)
(37,0.65)
(38,0.64374)
(39,0.65176)
(40,0.63856)
(41,0.65666)
(42,0.65034)
    };
    \addlegendentry{\small{AE}}

    \addplot[smooth,orange!70,mark=*,  error bars/.cd,
    y dir=both,
    y explicit,
] plot coordinates {
(1,0.5)
(2,0.5)
(3,0.509200)
(4,0.50300)
(5,0.52700)
(6,0.537800)
(7,0.565320)
(8,0.593620)
(9,0.60140)
(10,0.61120)
(11,0.624400)
(12,0.62800)
(13,0.62200)
(14,0.616300)
(15,0.624720)
(16,0.6442)
(17,0.61944)
(18,0.6332)
(19,0.6640)
(20,0.6542)
(21,0.6522)
(22,0.65136)
(23,0.6672)
(24,0.6764)
(25,0.6408)
(26,0.679)
(27,0.65924)
(28,0.65666)
(29,0.6368)
(30,0.658)
(31,0.6386)
(32,0.6534)
(33,0.6346)
(34,0.64978)
(35,0.62762)
(36,0.6426)
(37,0.64636)
(38,0.626820)
(39,0.6522)
(40,0.6512)
(41,0.655)
(42,0.6388)

    };
    \addlegendentry{\small{No pretrain}}
    
    \end{axis}
    \end{tikzpicture}
    } & 
    \resizebox {0.43\textwidth} {!} {
\begin{tikzpicture}
    \begin{axis}[
	width=0.8\linewidth,
	height=0.55\linewidth,
        xlabel= {\small Epochs},
        xlabel style={yshift=0.2cm,xshift=-0.2cm},
	legend pos =  north west
]
    \addplot[smooth,color=blue,mark=square*,error bars/.cd,y dir=both,y explicit]
        plot coordinates {

(1,0.5112	)
(2,0.51178	)
(3,0.52118	)
(4,0.5164	)
(5,0.53488	)
(6,0.5694	)
(7,0.5426	)
(8,0.5896	)
(9,0.60036 	)
(10,0.5814	)
(11,0.6202	)
(12,0.5958	)
(13,0.6358	)
(14,0.62154	)
(15,0.6458	)
(16,0.65188	)
(17,0.6638	)
(18,0.6438	)
(19,0.651	)
(20,0.63826	)
(21,0.64678	)
(22,0.6748	)
(23,0.65	)
(24,0.659	)
(25,0.6518	)
(26,0.6724	)
(27,0.6748	)
(28,0.6976	)
(29,0.7074	)
(30,0.6628	)
(31,0.68622 )
(32,0.6942	)
(33,0.6944	)
(34,0.686	)
(35,0.643258)
(36,0.65688	)
(37,0.70102	)
(38,0.67842	)
(39,0.66644	)
(40,0.69586	)
(41,0.68072	)
(42,0.66554	)
        };
    \addlegendentry{\small{Ours (LSSL)}}
    \addplot[smooth,color=blue!40,mark=x,error bars/.cd,y dir=both,y explicit,]
        plot coordinates {
(1, 0.51022)
(2, 0.51584)
(3, 0.51954)
(4, 0.58032)
(5, 0.58714)
(6, 0.62312)
(7, 0.61162)
(8, 0.6141)
(9, 0.634094)
(10,0.58352)
(11,0.6208)
(12,0.6476)
(13,0.658)
(14,0.6296)
(15,0.5624)
(16,0.622)
(17,0.64472)
(18,0.6512)
(19,0.6588)
(20,0.5954)
(21,0.60572)
(22,0.6537)
(23,0.64476)
(24,0.6352)
(25,0.6028)
(26,0.603)
(27,0.6432)
(28,0.65082)
(29,0.646)
(30,0.64172)
(31,0.61794)
(32,0.63472)
(33,0.6332)
(34,0.66872)
(35,0.63262)
(36,0.6012)
(37,0.61576)
(38,0.6362)
(39,0.67286)
(40,0.6265	)
(41,0.62936	)
(42,0.65372	)
        };
    \addlegendentry{\small{SimCLR}}
    \addplot[smooth,color=green,mark=o,error bars/.cd,y dir=both,y explicit,]
        plot coordinates {
(1, 0.52562)
(2, 0.52722)
(3, 0.48768)
(4, 0.47682)
(5, 0.48452)
(6, 0.48822)
(7, 0.47242)
(8, 0.484694)
(9, 0.48028)
(10,0.49022)
(11,0.47782)
(12,0.48302)
(13,0.48282)
(14,0.4916)
(15,0.49082)
(16,0.48722)
(17,0.49262)
(18,0.48282)
(19,0.50222)
(20,0.50532)
(21,0.53348)
(22,0.49782)
(23,0.50794)
(24,0.49382)
(25,0.49582)
(26,0.50242)
(27,0.50842)
(28,0.51662)
(29,0.5126)
(30,0.514954)
(31,0.55264)
(32,0.5276)
(33,0.529)
(34,0.56768)
(35,0.571654)
(36,0.59202)
(37,0.58466)
(38,0.59716)
(39,0.58316)
(40,0.59786)
(41,0.5852)
(42,0.60478)
        };
    \addlegendentry{\small{$\beta$-VAE}}
    \addplot[smooth,color=red,mark=triangle*,error bars/.cd,y dir=both,y explicit,]
        plot coordinates {
(1, 0.5226)
(2, 0.56138)
(3, 0.5396)
(4, 0.54446)
(5, 0.53174)
(6, 0.5587)
(7, 0.556)
(8, 0.5566)
(9, 0.538)
(10,0.52152)
(11,0.55762)
(12,0.57146)
(13,0.547)
(14,0.5818)
(15,0.5584)
(16,0.5302)
(17,0.56006)
(18,0.54018)
(19,0.52868)
(20,0.5483)
(21,0.539154)
(22,0.547)
(23,0.55766)
(24,0.55854)
(25,0.57194)
(26,0.57434)
(27,0.56386)
(28,0.56086)
(29,0.57068)
(30,0.5926)
(31,0.57826)
(32,0.56386)
(33,0.57274)
(34,0.58774)
(35,0.5829)
(36,0.57706)
(37,0.5731)
(38,0.5742)
(39,0.5806)
(40,0.5872)
(41,0.5868)
(42,0.585)
        };
    \addlegendentry{\small{CPC}}
    \addplot[smooth,orange!70,mark=*,  error bars/.cd,
    y dir=both,
    y explicit,
] plot coordinates {

(1, 0.4794)
(2, 0.4794)
(3, 0.4794)
(4, 0.4766)
(5, 0.4744)
(6, 0.4744)
(7, 0.4744)
(8, 0.479)
(9, 0.479)
(10,0.4842)
(11,0.4858)
(12,0.50092)
(13,0.49752)
(14,0.50052)
(15,0.50332)
(16,0.509)
(17,0.5062)
(18,0.50752)
(19,0.50752)
(20,0.5066)
(21,0.5094)
(22,0.510764)
(23,0.51532)
(24,0.51412)
(25,0.51492)
(26,0.51372)
(27,0.51492)
(28,0.51412)
(29,0.51372)
(30,0.51372)
(31,0.51252)
(32,0.5168)
(33,0.5254)
(34,0.5274)
(35,0.5256)
(36,0.5262)
(37,0.5282)
(38,0.5258)
(39,0.5266)
(40,0.5278)
(41,0.52826)
(42,0.52826)
    };
    \addlegendentry{\small{No pretrain}}
    \addplot[smooth,black!70,mark=*,  error bars/.cd,
    y dir=both,
    y explicit,
] plot coordinates {
(1, 0.47974)
(2, 0.46674)
(3, 0.51122)
(4, 0.4876)
(5, 0.5146)
(6, 0.5146)
(7, 0.5202)
(8, 0.5299)
(9, 0.538)
(10, 0.5274)
(11, 0.5146)
(12, 0.519)
(13, 0.5146)
(14, 0.5298)
(15, 0.5274)
(16, 0.5146)
(17, 0.52334)
(18, 0.5146)
(19, 0.52084)
(20, 0.519)
(21, 0.5164)
(22, 0.5258)
(23, 0.536)
(24, 0.5266)
(25, 0.519)
(26, 0.5264)
(27, 0.5266)
(28, 0.5146)
(29, 0.519)
(30, 0.527)
(31, 0.527)
(32, 0.5146)
(33, 0.519)
(34, 0.536)
(35, 0.5254)
(36, 0.534286)
(37, 0.53396)
(38, 0.535536)
(39, 0.5326)
(40, 0.5354)
(41, 0.53022)
(42, 0.5354)

    };
    \addlegendentry{\small{AE}}
    \end{axis}
    \end{tikzpicture}
    } \\
    {\footnotesize (a) Single time point classification (CNN)} &
    {\footnotesize (b) Longitudinal classification (CNN+RNN)} \\ 
\end{tabular}
    \caption{Average accuracy in the first 45 epochs for classifying alcohol dependence.}
    \label{fig:lab_convergence}
\end{figure}
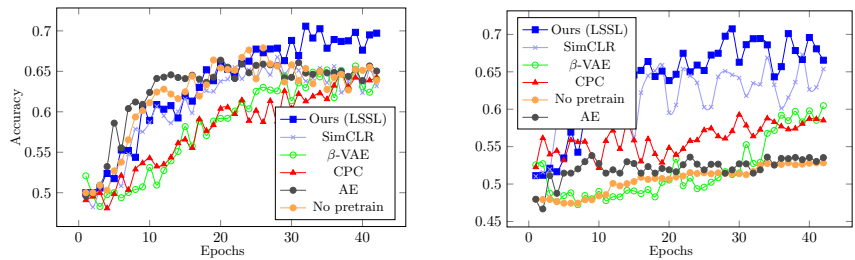

Prior literature indicates that alcohol dependence is only weakly separable from the control group \citep{Adeli2018multilabel,ouyang2020recurrent}, which is echoed in our results by the significantly lower \rev{bAcc} in the frozen setting (all methods $<$ 65\%, Table \ref{table:adni}) compared to the ADNI experiment. Nevertheless, all accuracy scores associated with LSSL were significantly better than chance based on the Fisher's exact test ($p<0.001$). Moreover, LSSL resulted in the fastest converging rate and highest accuracy upon convergence in the fine-tuning setting compared to the baselines (Fig. \ref{fig:lab_convergence}). The more challenging task of classifying alcohol dependence compared to the AD classification also took the RNN longer to converge for several baselines, which was not the case for LSSL.

\begin{table}[!t]
  \caption{Alcohol Dataset: \rev{Balanced accuracy (bAcc) of} cross-sectional and longitudinal classification  with and w/o fine-tuning the encoder. Best result in each column has bold typeset, and the second best is underlined.}
  \label{table:aud}
  \centering
  \begin{tabular}{l c c }
    \toprule
    & \multicolumn{2}{c}{Alcohol Dependence} \\
    \cline{2-3} 
    Pre-training & CNN & CNN+RNN       \\
    \cline{2-3} 
    Model     & frozen/fine-tuned & frozen/fine-tuned \\
    \hline
    No pretrain &  - / 69.5 & - / 52.8\\
    AE & 58.8 / 69.1 & 52.1 / 53.2 \\
    VAE \citep{kingma2013autoencoding} & 55.4 / 70.2 & 63.0 / 65.6\\
    $\beta$-VAE \citep{Higgins2017betaVAELB} & 52.1 / 67.5 & 60.8 / 61.0 \\
    SimCLR \citep{simclr} & \textbf{63.2} / \underline{68.7} & \underline{66.2} / \underline{69.3} \\
    CPC \citep{oord2018representation} & 51.9 / 67.5 & 62.0 / 63.2  \\
    Ours (LSSL) &  \underline{62.9} / \textbf{71.7} & \textbf{67.0} / \textbf{72.0}  \\
    \bottomrule
  \end{tabular}
\end{table}

\section{Discussion}
In recent years, many studies have used supervised models to estimate brain age from structural MRIs. These models have to be trained on a healthy population to predict chronological age to establish the normal association between brain age and brain structures \citep{Kaufmann2019,Smith2020eLife,zhao2019variational}. Then the trained model is applied to a diseased cohort to examine the brain age gap (difference between the estimated brain age and chronological age) induced by the disease \citep{Kaufmann2019}. This gap, however, is likely to be biased by the `domain shift' between the training and testing sets. LSSL alleviates this issue by training on both control and diseased subjects (without using their labels), which results in an analysis impartial to cohort bias. Moreover, the chronological age is a sub-optimal ground-truth for brain age in supervised models as there are multiple modes of brain aging within the healthy population due to genetic influence \citep{Smith2020eLife} (see also Fig. \ref{fig:age_vs_brainage}). LSSL addresses this challenge by omitting the supervision of chronological age and only using ordinal information of within-subject scans, which permits the characterization of heterogeneity across subjects. 

The advantage of LSSL over other unsupervised/self-supervised baselines is evident from our post-hoc classification, which revealed LSSL could learn discriminative cues within the representations even without using the group labels for training. However, the AD classification of LSSL did not rival with the highest accuracy score reported on the ADNI dataset \citep{Liu2017adni,Liu2019adni}, which was expected as we refrained from extensively exploring network architectures for the encoder. This type of exploration is a research direction orthogonal to the proposed self-supervised learning strategy in the representation space. This self-supervised learning strategy outperformed the baselines, all of which used  the same encoder setting. The setting was based on the most basic components used in standard CNNs (convolution, ReLU, and max-pooling). Therefore, we expect the findings revealed herein are likely to generalize to more advanced encoder architectures. 

\rev{\textbf{Limitation.}} \rev{As longitudinal studies are typically designed to examine the influence from time-dependent factors, we did not model \textit{Condition 2}, i.e., the independence between $\psi$ and other static factors, such as gender. This theoretically makes LSSL only result in a necessary condition for disentanglement.  In practice, we can examine whether \textit{Condition 2} holds via post-hoc analyses.} For example, in both experiments, the speed of brain aging was not significantly different between males and females ($p>0.05$ two-sided two-sample $t$-test) indicating an exact disentanglement between brain age and gender. 

\rev{Another limitation of LSSL} is that it does not eliminate the possible confound by other time-evolving factors co-occurring with brain age. For example, the change of body mass index between visits can also alter the overall brain volume and thereby the image representation \citep{Ward2005}. \rev{Although our pipeline could reduce} the impact of the change in brain volume by affinely registering each scan of the longitudinal MR images to a template, \rev{a systematic way of modeling confounders during the training of LSSL needs to be further explored. }

\rev{Finally, the current formulation of LSSL can only disentangle one time-variant factor related to brain age. This constraint limited our analysis to focus on abnormalities in brain age due to a neurological condition (such as AD). To model other disease effects not related to accelerated aging, a future research direction of LSSL is to jointly disentangle two orthogonal directions in the latent space so that one could separately characterize disease progression and brain aging.  }

\section{Conclusion}
In this work, we proposed a self-supervised representation learning framework called LSSL that incorporated theoretical advantages from the repeated measures design in longitudinal neuroimaging studies. The explicit longitudinal self-supervision permitted separate definitions for the factor and representation spaces, thereby omitting the ambiguity often encountered in fully unsupervised disentanglement models. Based on optimizing the colinearity between a global direction in the representation space and the developmental trajectories from subject-specific image pairs, LSSL successfully disentangled the factor of brain aging in the representation space, which was used to characterize normal aging pattern across the life span and to reveal the accelerated aging effects of Alzheimer's Disease and alcohol dependence. Compared to several other state-of-the-art representation learning methods, the pre-trained encoder and representations learned by LSSL are more suitable for supervised classification of diagnosis labels in various settings, indicated by faster convergence and higher (or equally high) prediction accuracy upon convergence.

\section*{Acknowledgement}
This work was supported by NIH Grants MH113406,
AA005965, AA010723, and AA017347, and by Stanford HAI AWS Cloud Credit. 
\bibliographystyle{model2-names.bst}\biboptions{authoryear}
\bibliography{refs}

\end{document}